\documentclass[fleqn,10pt]{wlscirep}
\usepackage[utf8]{inputenc}
\usepackage[T1]{fontenc}

\usepackage{subcaption}
\usepackage{physics}
\usepackage[printonlyused,nohyperlinks]{acronym}
\usepackage{soul}
\usepackage{multirow}
\newacro{PANN}{Physics Aware Neuromorphic Network}
\newacro{EO}{Earth Observation}
\newacro{AUPRC}{Area Under the Precision—Recall Curve}
\newacro{ML}{Machine Learning}
\newacro{UMAP}{Uniform Manifold Approximation and Projection}

\title{Training--free AI for Earth Observation Change Detection using Physics Aware Neuromorphic Networks}

\author[1]{Stephen Smith}
\author[1,2,3]{Cormac Purcell}
\author[1,4,*]{Zdenka Kuncic}
\affil[1]{University of Sydney, School of Physics, Sydney, Australia}
\affil[2]{University of New South Wales, School of Computer Science, Sydney, Australia}
\affil[3]{Trillium Technologies Pty Ltd., Australia}
\affil[4]{Emergentia Pty Ltd., Australia}
\affil[*]{zdenka.kuncic@sydney.edu.au}


\begin{abstract}
Earth observations from low Earth orbit satellites provide vital information for decision makers to better manage time--sensitive events such as natural disasters. For the data to be most effective for first responders, low latency is required between data capture and its arrival to decision makers. A major bottleneck is in the bandwidth--limited downlinking of the data from satellites to ground stations. One approach to overcome this challenge is to process at least some of the data on-board and prioritise pertinent data to be downlinked. In this study we propose a \ac{PANN} to detect changes caused by natural disasters from a sequence of multi-spectral satellite images and produce a change map, enabling relevant data to be prioritised for downlinking.
The \ac{PANN} used in this study is motivated by physical neural networks comprised of nano-electronic circuit elements known as ``memristors'' (nonlinear resistors with memory).  The weights in the network are dynamic and update in response to varying input signals according to memristor equations of state and electrical circuit conservation laws. The \ac{PANN} thus generates physics--constrained dynamical output features which are used to detect changes in a natural disaster detection task by applying a distance-based metric. Importantly, this makes the whole model training--free, allowing it to be implemented with minimal computing resources. The \ac{PANN} was benchmarked against a state-of-the-art AI model and achieved comparable or better results in each natural disaster category.  
It thus presents a promising solution to the challenge of resource--constrained on-board processing.
\end{abstract}
\begin{document}

\flushbottom
\maketitle
%
%
\thispagestyle{empty}

\section*{Introduction}
\acresetall{}

\ac{EO} from low Earth orbit satellites provide large volumes of data which are used in a diverse number of applications, including land cover monitoring \cite{vali2020deep}, food security \cite{maguluri2024sustainable} and gas leaks \cite{ruuvzivcka2023semantic}. \ac{ML} techniques have become a critical tool for processing the enormous quantity of \ac{EO} data \cite{ferreira2020monitoring, parelius2023review}, with new \ac{ML} models being designed for change detection tasks that specifically use \ac{EO} data \cite{diakogiannis2021looking, chen2020spatial}.
An area in which \ac{ML} using \ac{EO} data can have a significant impact is the management of natural disasters \cite{bello2014satellite, huyck2014remote, le2020space}. For the data to be most effective for first responders, low latency is required between data capture and its arrival to decision makers. One of the largest bottlenecks is the downlinking of data due to the limited bandwidth and coordination of available ground stations \cite{karapetyan2015satellite}. This problem is expected to grow in the future as more data is collected on board satellites due to advances in sensor technology (e.g. hyper-spectral cameras) and an increasing number of satellites \cite{selva2012survey}. One way to overcome this challenge is to process some of the data on-board and prioritise pertinent data to be downlinked, particularly for time-sensitive applications like natural disaster management. On-board processing will also be critical for future spacecraft deployed to remote sites in the solar system, or for monitoring solar weather in near-real-time.

Traditional algorithms have been developed to detect changes in \ac{EO} images, for example BFAST \cite{verbesselt2010detecting}, which has been applied to detecting changes in forests \cite{rodriguez2024monitoring}. Algorithms such as BFAST rely on spectral indices as input. These are a relatively simple pixel comparison method to highlight objects of interest such as burn areas \cite{normalizedBurnRatio}, diseased crops \cite{zheng2018new} and flood mapping \cite{farhadi2025introducing}. These indices are effective as they utilise domain knowledge regarding the object of interest. For example, the normalised difference vegetation index is commonly used to incorporate knowledge about the wavelengths that the Chlorophyll pigment absorbs \cite{pettorelli2013ndviBook}. More sophisticated indices, which involve more than just a normalised difference between two spectral bands, have been proposed to improve the detection \cite{normalizedBurnRatio, jiang2008development}. The advantage of spectral indices is that they provide a simple, low computational cost method to determine objects of interest quickly, with reasonable accuracy. Additionally, they do not require training or large datasets to deploy, unlike \ac{ML} models.

AI models based on deep \ac{ML}, specifically Convolutional Neural Networks (CNNs), have been applied to several natural disaster tasks using \ac{EO} data, including flood segmentation \cite{nemni2020fully, rudner2019multi3net}, detecting earthquake affected buildings \cite{qing2022operational} and volcanic eruptions \cite{del2021board}. To increase the performance of these AI models, often the number of layers and weights within the model needs to be increased. Chintalapati \textit{et al.} \cite{chintalapati2024opportunitiesOnboardAI} showed how the architecture of CNN models has evolved to increase performance on an \ac{EO} image classification task, which resulted in an increase in the number of weights in the models, with the two best performing models, CoCa \cite{yu2022coca} and BASIC-L \cite{chen2023symbolic}, using 2.1 and 2.44 billion parameters, respectively. It was noted that most deep learning AI models have so many parameters that they would exceed the memory budget on most \ac{EO} satellites. In recent years, work has been done to design smaller AI models specifically for use on-board satellites \cite{abid2021ucl, ravaen}. Furthermore, missions such as $\Phi$-Sat-1\cite{giuffrida2021varphi} and OPS-SAT \cite{labreche2022ops} have also designed satellites to accommodate AI models on-board. Although these models have been specifically designed for on-board use, all the training still occurs on the ground because of the large computational overhead. Only inference is performed on-board. Since AI models are trained on the ground, they often cannot be trained with the on-board sensor data, leading to degraded performance \cite{koh2021wilds, derksen2021fewShotChallenge}. Work has been done to retrain models, once operational, with new data as it is acquired \cite{mateo2023inorbitTraining, ruuvzivcka2023fast}.

In this study, we introduce the \ac{PANN}, a AI model with a novel architecture modelled after physical neural networks comprised of nano-electronic circuit elements known as memristors (nonlinear resistors with memory) that mimic synapses \cite{christensen20222022, song2023recent, Kuncic-Nakayama_2021}. Like biological synapses, the memristive network weights continuously update in response to varying input signals, constrained by physical laws embedded within the model. The \ac{PANN}, therefore, actively learns as new inputs are fed into the network, such as from a sensor on-board a satellite. As such, the \ac{PANN} takes a significant shift away from the conventional artificial neural network AI paradigm, as the gradient--descent, back--propagation training of weights is completely replaced by physics--constrained dynamical equations. 
This is also in contrast to physics--informed neural networks, where the physics is embedded as a component of the model in the artificial neural network paradigm. Typically, this is done by modifying activation functions, gradient descent optimisation techniques, network structure, or loss functions \cite{karniadakis2021physics,cuomo2022scientific}. The added physics normally takes the form of a partial differential equation and is task specific, for example, incompressible Navier-Stokes equations for fluid flow problems \cite{jin2021nsfnets}, whereas with the \ac{PANN}, the physics is independent of the task.

Physical memristive networks, which the \ac{PANN} is modelled after, have demonstrated learning tasks such as image classification and sequence memory \cite{Zhu2023online}, voice recognition \cite{Lilak2021,Kotooka2024}, long-/short-term memory and contrastive learning \cite{Loeffler2023neuromorphic}, as well as regression tasks of varying complexity \cite{Sillin2013theoretical,Stieg2014_self-org,Michieletti2025soc} and chaotic time series prediction \cite{Milano2022materia}.
The PANN model has also been used to demonstrate these and other ML tasks, including transfer learning and multi-task learning \cite{kuncic2020neuromorphic,Fu2020,hochstetter2021avalanches,zhu2020harnessing,zhu2021information,Loeffler2021,Zhu2023_L2L,Baccetti2024}. These \ac{ML} tasks were implemented using reservoir computing, where only weights in a single output layer need to be trained.
This approach offers a substantial advantage over deep neural network models as it does not need computationally intensive training algorithms and thus does not consume as much energy. These properties make the \ac{PANN} an ideal candidate for use on-board satellites.  

In this study, we use a natural disaster dataset released by R{\r{u}}{\v{z}}i{\v{c}}ka \textit{et al.} \cite{ravaen} to evaluate the performance of the \ac{PANN} model on a change detection problem, that contains a time series of multi-spectral \ac{EO} images from Sentinel-2. We benchmark the \ac{PANN} against a state-of-the-art AI model and achieve comparable or better results. 
We also evaluate the performance of the \ac{PANN} on a secondary dataset of the same natural disasters using multi-spectral \ac{EO} images from LandSat-8, to compare the models performace on different sensors.
Additionally, we use a distance measurement to detect changes, bypassing the need for an output layer altogether. Hence, we do not need to train any part of the model and the \ac{PANN} does not need to be trained on the ground prior to deployment.

This article is organised as follows: in ``\hyperref[sec:Results]{Results}'' we show the results from the \ac{PANN} model and compare them to a deep learning model for the same change detection task. We also show visualisations of the features the \ac{PANN} model extracts. In ``\hyperref[sec:Discussion]{Discussion}'' we outline key findings from the results and their implication for use on-board satellites. Finally, ``\hyperref[sec:Methods]{Methods}'' presents the details of the \ac{PANN} model and workflow.

\section*{Results}
\label{sec:Results}

The \ac{PANN} model is used here to detect regions of change from natural disasters using \ac{EO} multi-spectral images. The change detection task requires the model to take in a sequence of five images, where the final image is after a natural disaster event. The model outputs a change feature map that highlights the regions affected by the natural disaster, by breaking the images into smaller tiles and assigning a change likelihood score for each tile. The \ac{PANN} model used in this study is an adaptive network, with weights that adjust with each new input. This allows the \ac{PANN} to generate features from the inputs without requiring any training (see \hyperref[sec:Methods]{Methods} for details). To determine if a tile has changed, the distance between inputs are compared in the feature space rather than the input space (pixel comparison). By applying a distance-based metric directly to the feature space, no training is performed at any stage in the model. The dataset includes four different categories of natural disaster events: fires, floods, hurricanes and landslides. Additionally, each natural disaster event is accompanied by a change mask which is used only to evaluate the performance of the \ac{PANN} model.

\begin{figure}
\centering
\begin{subfigure}{\linewidth}
\includegraphics[width=\linewidth]{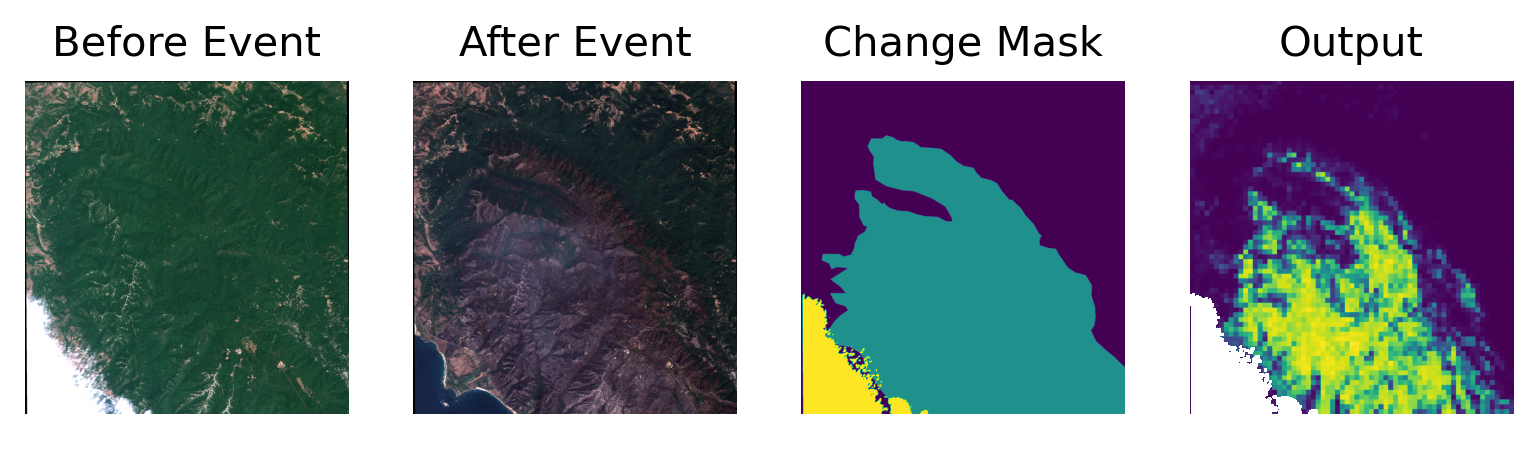}
\end{subfigure}
\begin{subfigure}{\linewidth}
\centering
\includegraphics[width=\linewidth]{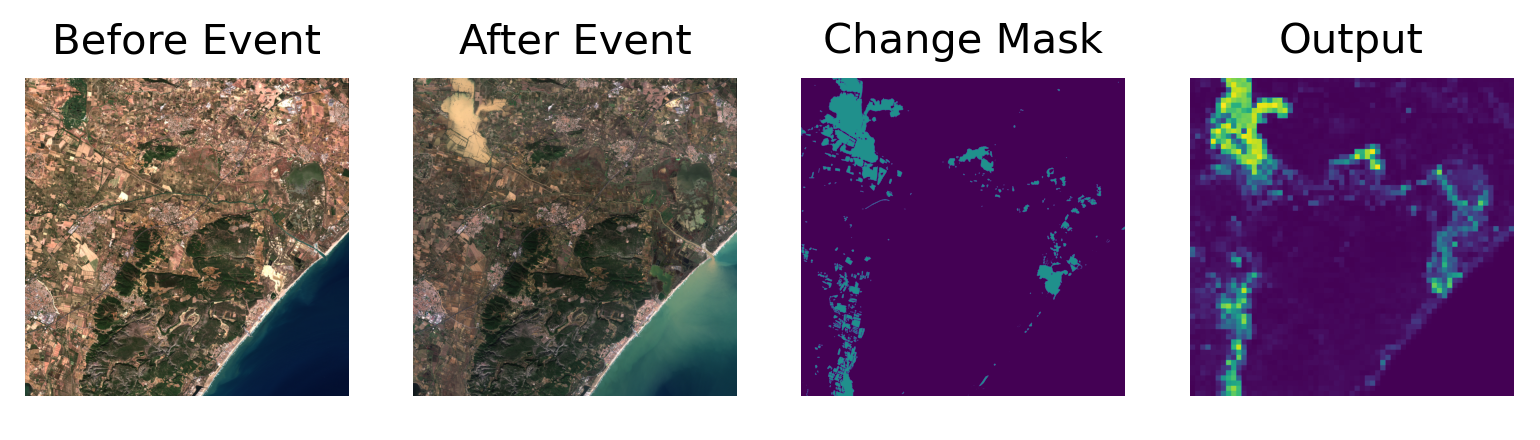}
\end{subfigure}
\begin{subfigure}{\linewidth}
\centering
\includegraphics[width=\linewidth]{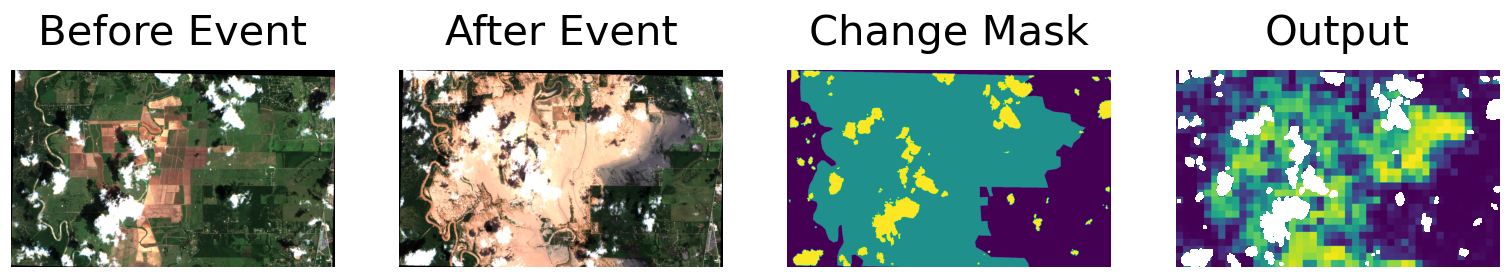}
\end{subfigure}
\begin{subfigure}{\linewidth}
\centering
\includegraphics[width=\linewidth]{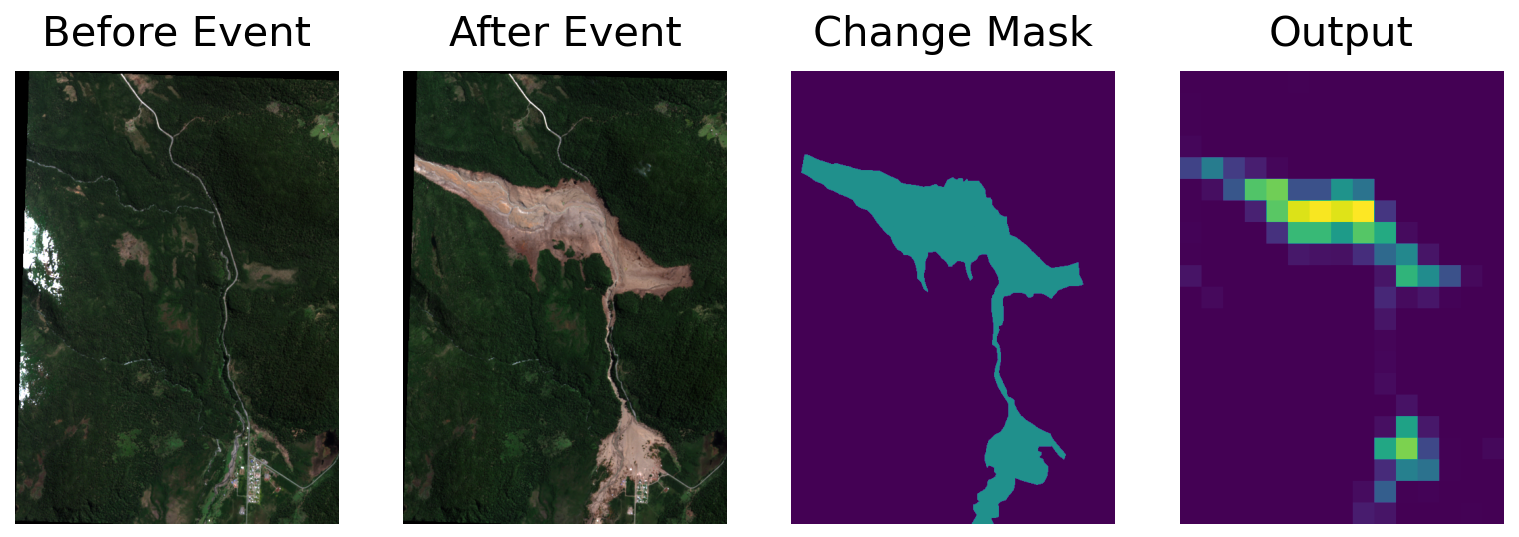}
\end{subfigure}
\caption{RGB images (columns 1 and 2) directly before and after a natural disaster event, the corresponding target change mask (column 3) and the output change map produced by the \ac{PANN} model (column 4) for each type of natural disaster: Top row -- an example fire natural disaster event; Second row -- an example flood natural disaster event; Third row -- an example hurricane natural disaster event; Bottom row -- an example landslide natural disaster event. Cloud-cover has been masked from the images and corresponds to yellow in the change mask, with genuine differences coloured aqua. Satellite images are from Sentinel-2 complied by R{\r{u}}{\v{z}}i{\v{c}}ka \textit{et al.} and available in their git\cite{ravaenData}.}
\label{fig:changeMaps}
\end{figure}

Figure~\ref{fig:changeMaps} shows the output change maps produced by the \ac{PANN} model for each type of natural disaster event. Alongside the change maps are the RGB images directly before and after the natural disaster events and the associated change mask, which also includes clouds from the images directly before and after the natural disaster event. Although only one image is shown before the natural disaster event in Fig.~\ref{fig:changeMaps}, a total of four images before the natural disaster event were used to create the displayed change maps. For the four natural disaster events, the main areas of change have been correctly identified as changes in the change maps, with a limited amount of noise in the images being considered as a change, such as the vegetation colour change in the fire natural disaster event (top row), or the sediment being washed into the ocean in the flood natural disaster event (second row). 
Clouds are detected as changes, but are masked out of the change maps shown in Fig.~\ref{fig:changeMaps}, since they are ignored when evaluating the change maps to allow for a more direct comparison to a state-of-the-art deep learning model (see \hyperref[sec:Methods]{Methods} for details). 

For each natural disaster event a change map was created, with the order in which the natural disasters were passed into the \ac{PANN} model being randomised. We evaluated the quality of the change maps by calculating the \ac{AUPRC}, which we create tile--wise. The change likelihood score for each tile is given by equation~(\ref{eqn:changeScore}) in \hyperref[sec:Methods]{Methods}, which depends on the distance metric used to measure the position of the tiles in the feature space. Table~\ref{tab:distMetric} reports the overall \ac{AUPRC} as a percentage for each of the natural disaster categories using three different distance metrics, Euclidean, Cosine and Correlation distances. The Correlation distance significantly outperforms the other distance metrics and is also much more consistent across the different natural disaster classes.

\begin{table}
\centering
\begin{tabular}{|l|l|l|l|l|}
\hline
Distance Metric & Fires & Floods & Hurricanes & Landslides \\
\hline
Euclidean & $90.084 \pm 0.007$ & $42.79 \pm 0.04$ & $43.1 \pm 0.07$ & $27.38 \pm 0.09$ \\
\hline
Cosine & $86.71 \pm 0.01$ & $39.36 \pm 0.03$ & $22.50 \pm 0.01$ & $35.2 \pm 0.2$ \\
\hline
Correlation & $\mathbf{93.66 \pm 0.01}$ & $\mathbf{59.26 \pm 0.02}$ & $\mathbf{76.51 \pm 0.05}$ & $\mathbf{73.5 \pm 0.2}$ \\
\hline
\end{tabular}
\caption{Comparison between different distance metrics used to calculate the \ac{PANN} change score for different natural disaster classes. All values are calculated as mean \ac{AUPRC} (\%) $\pm$SEM over 5 runs. Best scores are highlighted.}
\label{tab:distMetric}
\end{table}

We benchmark our \ac{PANN} model against a state-of-the-art variational autoencoder model \cite{kingma2013VAE}, called RaVAEn, which was specifically designed to be a relatively small AI model that could be deployed on-board satellites. We also compare the \ac{PANN} model to a baseline method that uses the same change score equation but compares the different tiles in the pixel space rather than the feature space like the other two AI models. 
When comparing models, it should be noted that the \ac{PANN} has an automatic feature engineering process incorporated into the model setup that is not present in the RaVAEn model or the baseline method.
Table~\ref{tab:mainResults} reports the performance of the three different models for each of the natural disaster classes. Both AI models, RaVAEn and \ac{PANN}, outperform the baseline method in all classes. The RaVAEn and \ac{PANN} models achieve comparable results in the fires, hurricanes and landslides, with the \ac{PANN} slightly ahead for the fires and hurricanes, while RaVAEn is slightly ahead for the landslides. With the flood natural disasters, the \ac{PANN} achieves the highest results by a significant margin with a score 14\% higher than that of RaVAEn.

\begin{table}
\centering
\begin{tabular}{|l|l|l|l|l|}
\hline
Model & Fires & Floods & Hurricanes & Landslides \\
\hline
Baseline & 86.5 & 37.6 & 58.0 & 62.9 \\
\hline
RaVAEn & $90.98 \pm 0.08$ & $44.5 \pm 0.7$ & $76.10 \pm 0.08$ & $\mathbf{76.5 \pm 0.8}$ \\
\hline
PANN & $\mathbf{93.66 \pm 0.01}$ & $\mathbf{59.26 \pm 0.02}$ & $\mathbf{76.51 \pm 0.05}$ & $73.5 \pm 0.2$ \\
\hline
\end{tabular}
\caption{Comparison between change scores predicted by different models for different natural disaster classes. All values are calculated as \ac{AUPRC} (\%) $\pm$SEM over 5 runs, with values for the \ac{PANN} model corresponding to the best scores from Table~\ref{tab:distMetric}.}
\label{tab:mainResults}
\end{table}

To assess how well the \ac{PANN} model generalises to data from different sensors, we evaluated the model on a subset of the natural disaster events using \ac{EO} images from LandSat-8 and compared the results to the Sentinel-2 data (see \hyperref[sec:Methods]{Methods} for details). Table~~\ref{tab:landsatResults} reports the performance of the PANN and baseline models for the LandSat-8 data and the comparative results for the equivalent subset in the Sentinel-2 data. From the results it can be seen that the \ac{PANN} model outperforms the baseline method for both the data from the Sentinel-2 satellites and the LandSat-8 satellite for every natural disaster category. Both models achieve higher performance when using data from Sentinel-2 compared to LandSat-8. Additionally, the difference in performance between datasets is much larger for the baseline methods, which shows a significant decrease in performance when using the LandSat-8 data.

\begin{table}
\centering
\begin{tabular}{|l|l|l|l|l|l|l|}
\hline
\multirow{2}{*}{Model} & \multicolumn{2}{|c|}{Fires} & \multicolumn{2}{|c|}{Floods} & \multicolumn{2}{|c|}{Landslides} \\
\cline{2-7}
 & S2 & L8 & S2 & L8 & S2 & L8 \\
\hline
Baseline & 79.0 & 59.1 & 64.6 & 38.8 & 68.9 & 34.4 \\
\hline
PANN & $90.26\pm 0.02 $ & $86.07 \pm 0.02$ & $66.38 \pm 0.03$ & $53.9 \pm 0.1$ & $75.88 \pm 0.02$ & $53.3 \pm 0.2$ \\
\hline
\end{tabular}
\caption{Comparison between change scores for different sensors by natural disaster classes for the \ac{PANN} model and baseline method. S2: data from Sentinel-2 satellites, L8: data from LandSat-8 satellite. All values are calculated as \ac{AUPRC} (\%) $\pm$SEM over 5 runs. }
\label{tab:landsatResults}
\end{table}

\subsection*{Feature Space}

\begin{figure}
\centering
\includegraphics[width=\linewidth]{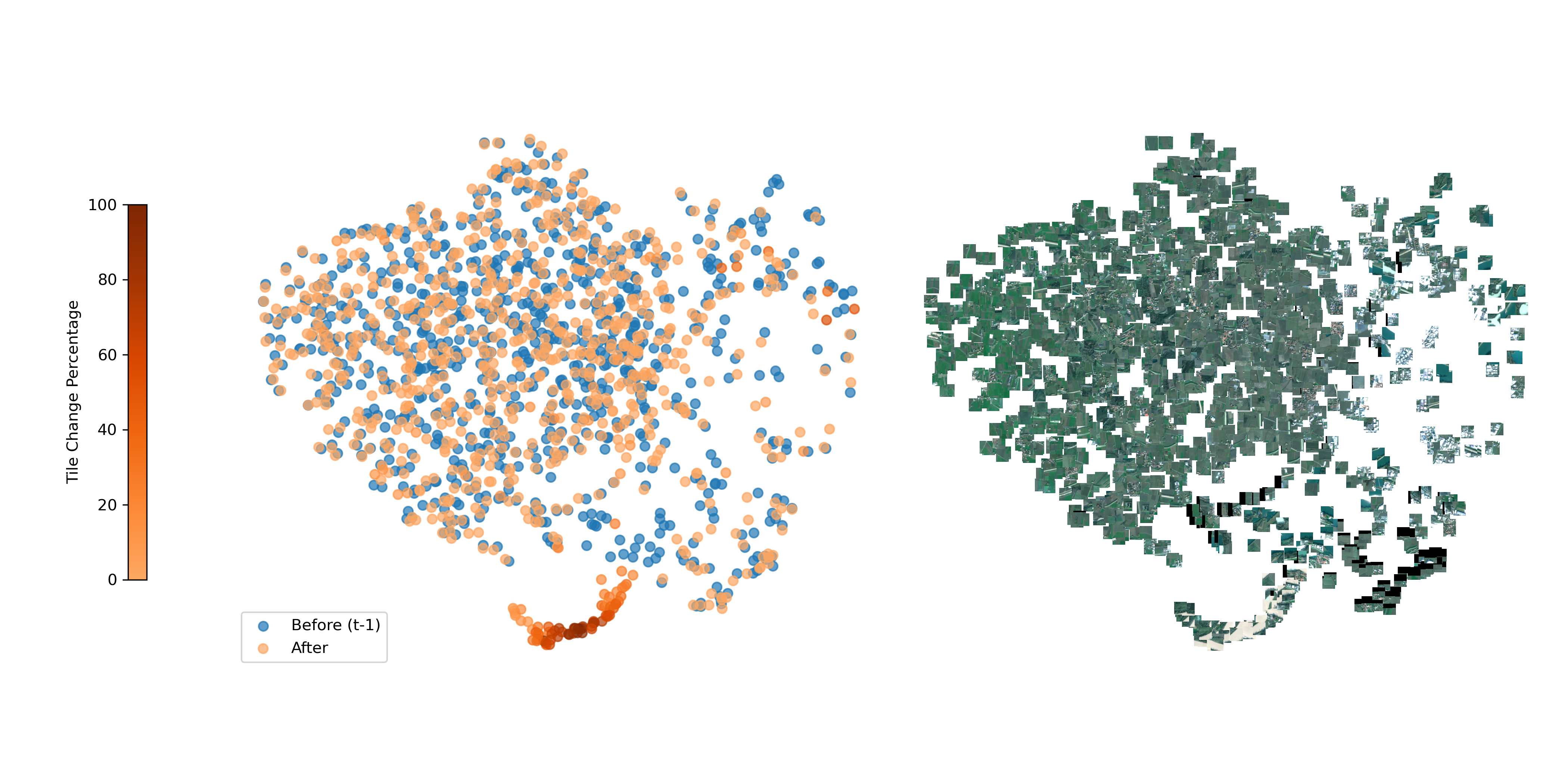}
\caption{2D \ac{UMAP} projections of the \ac{PANN} feature space, using the correlation distance for a flood natural disaster event: Left -- positions of before (blue) and after (orange) tiles, with percentage change in the after tiles indicated by the colour bar; Right -- the corresponding tile images. The tiles with flooding occupy a distinct arc-like cluster at the bottom of the projection.}
\label{fig:flood2D}
\end{figure}

The AI models RaVAEn and \ac{PANN} outperform the baseline method, since they can extract the most relevant features from the tiles and group similar tiles together, while keeping dissimilar tiles further apart in their feature spaces. To visualise the high dimensional feature space of the \ac{PANN}, we used the \ac{UMAP} \cite{mcinnes2018umap} method to reduce the dimensions of the feature space into 2D. The \ac{UMAP} used 20 neighbours, a minimum distance of 0.4 and the correlation distance, unless otherwise stated. Figure~\ref{fig:flood2D} shows the \ac{UMAP} 2D projection of the feature space for a flooding event. The left-hand side shows the position of the tiles in the image directly before and after the flooding event. For tiles after the flooding event, the percentage change in each tile is denoted by the colour bar. The right-hand side of the figure shows the same projection of the feature space but with the tile images instead of points. The change tiles are grouped together at the bottom of the projections and the remaining tiles are also sorted according to the features in the tiles, with greener farmland tiles on the left-hand side and tiles with the white objects on the right. Tiles that have missing values, from the north alignment of the swath, are also grouped towards the bottom right, with the tiles that have the missing values across the top being separated from the tiles with the missing values on the right-side of the tile. For comparison with the \ac{UMAP} projection for the pixel space, see Supplementary Fig. S1 online.

\begin{figure}
\centering
\includegraphics[width=\linewidth]{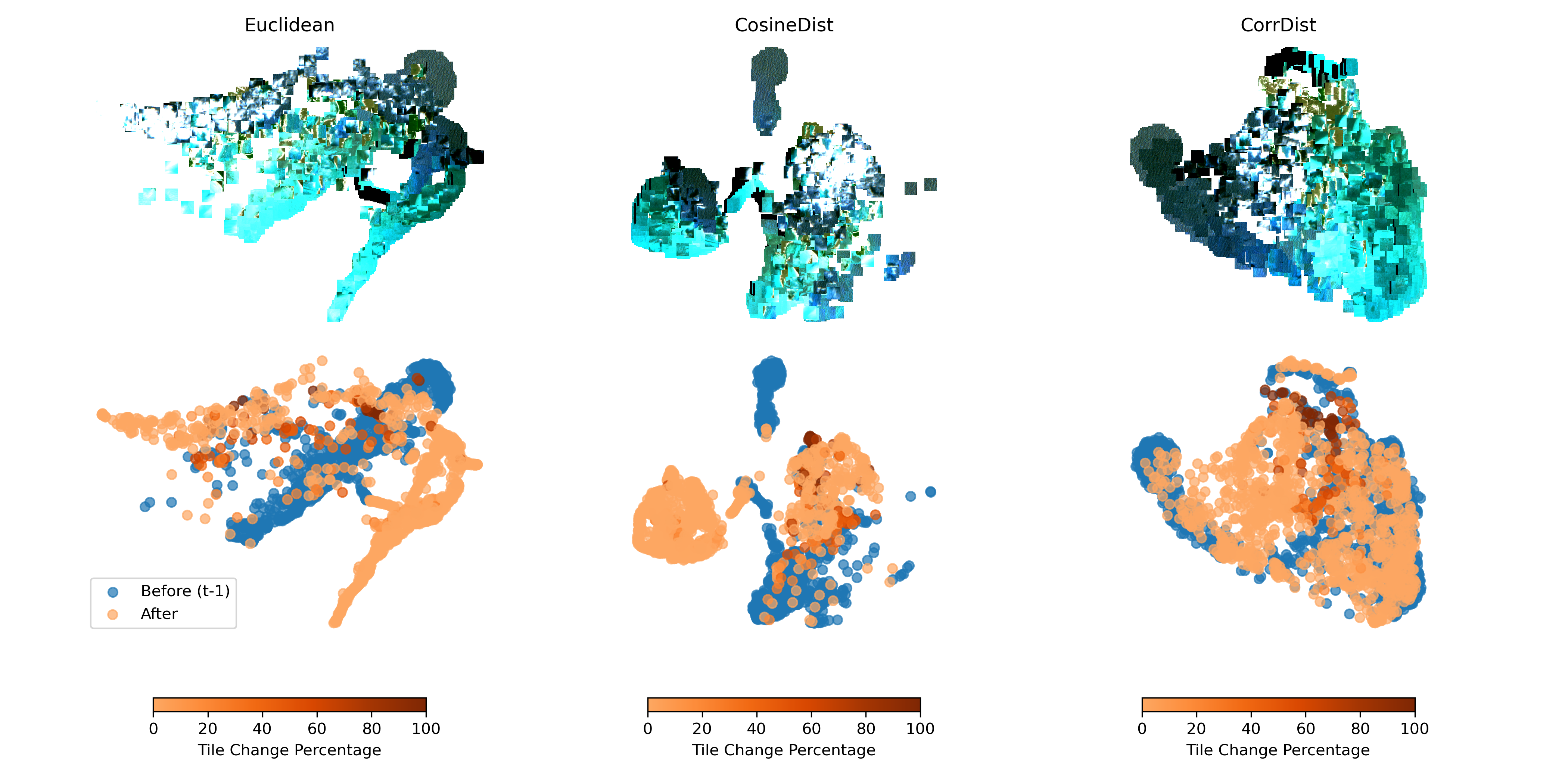}
\caption{Visual comparison between the 2D \ac{UMAP} projection of the \ac{PANN} feature space for a hurricane natural disaster using three different distance metrics: Euclidean, Cosine and Correlation distances.}
\label{fig:distanceFS}
\end{figure}

Like most dimensionality reduction techniques, \ac{UMAP} aims to reduce the number of dimensions while maintaining the overall structure and information of the original, higher dimensional space and thus depends on the distance metric used to measure the distances between points. Figure~\ref{fig:distanceFS} shows the \ac{UMAP} projection of the feature space for a hurricane natural disaster for the three distance metrics used in Table~\ref{tab:distMetric} and shows how the clustering of the tiles varies depending on the distance metric being used. This provides a visual qualitative tool to understand the large range in results reported in Table~\ref{tab:distMetric} due to the change in distance metric being used.

\section*{Discussion}
\label{sec:Discussion}

In this study, we introduced the \ac{PANN} model for a natural disaster change detection task and compared it against the state-of-the-art AI model RaVAEn and a baseline method. All three models receive the same multi-spectral \ac{EO} images for the inputs and use the same change detection equation given by equation~(\ref{eqn:changeScore}), for each tile in the image. The difference in the model performances, therefore, is attributed to their ability to extract the most relevant features from the inputs, while ignoring noise within the inputs. 
We note, however, that while the \ac{PANN} model receives the same inputs as the other models, only the inputs corresponding to the bands selected by the automatic feature engineering process, on a scene by scene basis, are passed to the relevant individual networks. This inherently allows the \ac{PANN} model to ignore irrelevant information as determined by the domain knowledge embedded in the feature engineering process. 
The baseline method does not extract any features from the inputs and instead does the comparison directly in the pixel space. This makes the baseline method more sensitive to noise in the images and, as expected, it exhibits the lowest performance as reported in Table~\ref{tab:mainResults}. The AI models, on the other hand, extract meaningful features from the tiles and hence outperform the baseline method in every natural disaster category. The \ac{PANN} and RaVAEn models achieve a comparable result in three of four of the natural disaster categories, demonstrating that both models extract equally useful features for these categories. For flood natural disasters, however, the \ac{PANN} model outperforms the RaVAEn model and thus arguably extracts more meaningful features. 

Although being able to extract the features is the most important task for the \ac{PANN} model, how the feature space is navigated is also important for the performance on the \ac{PANN} model. This is seen in Table~\ref{tab:distMetric} where the same feature space can produce significantly different results based on the distance metric used to navigate the feature space. This is visualised in the \ac{UMAP} projections in Fig.~\ref{fig:distanceFS}, where different distance metrics are used to create the projections. The feature space produced by the \ac{PANN} has a high dimensionality, as such the relative Euclidean distance between points tends to decrease\cite{xia2015effectivenessOfEuclidean}. This would result in the change likelihood score decreasing, particularly for change tiles and is likely why the Euclidean distance does not produce the best results.
What is surprising is the large difference between the Cosine and Correlation distances, given that both metrics are angle-based measures. The Cosine distance is given by equation~(\ref{eqn:cosineDist}) and measures the angle between the vectors in the feature space. The Correlation distance is the same as the Cosine distance except each sample is mean centred first, as given by equation~(\ref{eqn:corrDist}).
The mean-correction for the Correlation distance effectively changes the location of the origin and hence the angle between the vectors also changes \cite{korenius2007pcaCosineInformationRetriveal}. One possible reason why the mean correction has such a large impact on the performance is that it helps reduce covariate shift in the output sequence that can arise from the dynamical nature of the \ac{PANN}. The covariate shift effect can be seen in Fig.~\ref{fig:distanceFS} (and Supplementary Fig. S2), where there is a much greater overlap of points at the different timesteps (that are not changes) in the Correlation projection compared to that of the Cosine projection. Interestingly, batch normalisation layers were introduced to deep neural networks to help reduce the covariate shift between layers in the networks \cite{ioffe2015batchNorm} and have been widely accepted to speed up training times and increase performance. We note that the \ac{PANN} does not have any batch normalisation layers, while the RaVAEn model does, and that the best performing metrics for the models are the Correlation and Cosine distance, respectively.

\begin{figure}
\centering
\includegraphics[width=\linewidth]{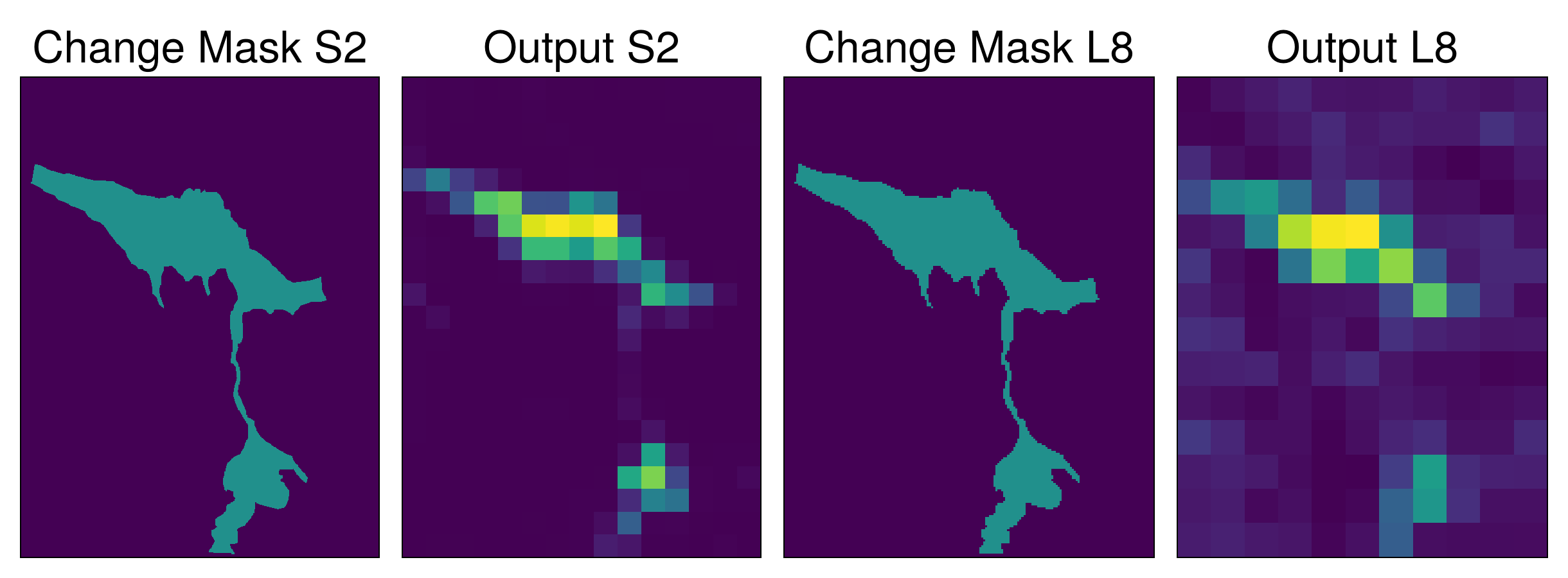}
\caption{Comparison between the change maps produce by the \ac{PANN} model using the Sentinel-2 (S2) and LandSat-8 (L8) dataset, for the same landslide natural disaster event. The corresponding change masks are also included.}
\label{fig:sensorComparison}
\end{figure}

From the results in Table~\ref{tab:landsatResults}, two clear trends are evident. The first is that both the \ac{PANN} and baseline methods perform better when using data from Sentinel-2 than LandSat-8 across all natural disaster categories. This can largely be explained by the decrease in resolution. The Sentinel-2 data has a spatial resolution of 10\,m, with a tile size $32 \times 32$ pixels and a tile  resolution of 320\,m. While the LandSat-8 data has a spatial resolution of 30 m, with a tile size of $16 \times 16$ pixels and tile resolution 480\,m. As a result, the models are not able to resolve finer details when determining if a change has occurred in a LandSat-8 tile. This difference in resolution can be seen in Fig.~\ref{fig:sensorComparison}, which shows the change mask and output change map for both datasets for one of the landslides. Additionally, in Fig.~\ref{fig:sensorComparison}, more noise is evident in the LandSat-8 change map, compared to the Sentinel-2 change map, particularly in the areas where there is no change. This would also contributes to the decrease in performance. The second trend that can be clearly seen in Table~\ref{tab:landsatResults}, is that the decrease in performance using the LandSat-8 data relative to the Sentinel-2 data is significantly less for the \ac{PANN} model compared to the baseline method. This shows that the \ac{PANN} model is able to generalise to different sensor data much better than the baseline method, further showing that the \ac{PANN} model is able to extract meaningful features from the input data.

Our results show that the PANN model achieves comparable-to-better results than the RaVAEn state-of-the-art AI model for the natural disaster change detection task conducted in this study. Most importantly, the PANN was able to achieve this result by creating a mapping from the input space to the feature space without any training. The PANN is training-free as it continuously adapts with each new input. The model equations are relatively light–weight and thus all the processing of the data by the PANN model was able to be carried out using only CPUs in this study. The compute resources, time and memory, were measured for the PANN to process a patch sequence of $574 \times 509$ pixels over 5 frames, corresponding to roughly a 25 km$^2$ region. Using 8 cores on an Intel Xeon Scalable processor, the PANN model takes 19\,s and uses approximately 1\,GB of memory (see Supplementary Fig.~S3). We note that the model could be optimised to further decrease memory requirements by reducing the floating-point precision of the PANN from 64 bit to 32 or 16 bit precision. We also note that the processing time for the PANN model is not directly comparable to the inference time of RaVAEn or other AI change detection models, as the processing time to evaluate a scene also includes updating of the PANN weights with each input. Unlike AI models, the PANN model does not require any additional time and compute resources for training. The physics–based equations that constitute the PANN model are a significant deviation from conventional deep multi-layered bipartite neural networks that typically use mathematical activation functions to perform nonlinear transformations and stochastic gradient decent with backpropagation as the learning algorithm to train networks weights into an optimal static state. As such, the iterative training process is computationally expensive and usually requires cloud-based GPUs \cite{thompson2020computationalLimits}, to produce a trained model. As the PANN simulates a physical device, based on memristive nanowire networks, we expect the computational resources to significantly decrease when tasks are implemented on the physical hardware (see e.g. \cite{Zhu2023online,milano2023materia,Loeffler2023neuromorphic,Milano2022materia,hochstetter2021avalanches}).

The \ac{PANN} model is relatively sensor agnostic since it does not need to be pre-trained with data from a specific sensor and the model design of having a separate network for each spectral band (see \hyperref[sec:Methods]{Methods} for details), allows it to receive images with varying number of spectral bands. Having a separate PANN for each band also allows a simple training–free method to add domain knowledge to the model in the form of an automatic feature engineering mechanism that determines which spectral bands and hence \acp{PANN} are used (see \hyperref[sec:Methods]{Methods} for details). In this study, only relatively minor adjustments to the feature engineering mechanism were required when changing the sensor data from the Sentinel-2 satellite to the LandSat-8 satellite. In addition, having a separate \ac{PANN} for each band combined with the feature engineering means only a subset of the \acp{PANN} are required, further reducing the computational resources. Having multiple smaller networks, however, means that interactions between bands may not be fully captured, as compared to a single larger network, where information from all the bands is integrated while being processed. Alternatively, another model design that could be considered is multiple networks with a hierarchical structure\cite{fan2015hierarchical}.

The training-free nature of \ac{PANN} makes it an ideal candidate for extreme edge computing applications, particularly for \ac{EO} analysis on-board satellites. Specifically, there are two main challenges that are inherently addressed by the training-free nature of the \ac{PANN}. The first is that \ac{ML} models typically do not generalise well and need to be trained for a specific sensor. When \ac{ML} models are trained using data from a different sensor, there is often a notable decrease in performance \cite{koh2021wilds, derksen2021fewShotChallenge}. Furthermore, training a \ac{ML} model with only simulation data can be difficult \cite{chintalapati2024opportunitiesOnboardAI}. The training-free nature of the \ac{PANN} therefore makes it a promising candidate to circumvent this issue. The second is that \ac{ML} models are often retrained as new data from the desired sensor become available to increase performance and to account for any data-shift problems. This process requires the new data to be downlinked, to retrain the model on the ground before uplinking the updated model to the satellite \cite{mateo2023inorbitTraining}. The dynamical nature of the \ac{PANN} would allow this whole process to be bypassed as the \ac{PANN} inherently updates and learns from each new input from the sensor.

In this study, we have demonstrated the feasibility of the \ac{PANN} for detecting changes from natural disasters within the context of their potential deployment on-board satellites. Based on these simulation results, future work would include testing a prototype of the hardware device, based on memristive nanowire networks, which the \ac{PANN} is modelled after. Based on previous studies with such neuromorphic devices~\cite{Zhu2023online,milano2023materia,Loeffler2023neuromorphic,Milano2022materia,hochstetter2021avalanches}, it is expected that implementation on the low-power hardware device will substantially reduce processing times and energy consumption, and could be achieved using only the on-board compute. The methodology outlined in this study could be used when implementing in hardware with one key change: the physical network devices may need larger electrode grids than the $16 \times 16$ grid used here, as additional electrodes would be needed to read out voltages from the larger physical networks. To date, devices with up to 128~electrodes have been fabricated~\cite{demis2016nanoarchitectonic}, although interestingly, experimental results suggest that using readouts from only a subset of electrodes may be sufficient to capture the most salient output features~\cite{Zhu2023online}. This may be attributed to how the electrodes integrate signals from many nanowires they make electrical contact with. This means that when implementing in a physical device, the option of using one large network that processes all the spectral bands together may be viable. Multiple configurations for the electrode placement would be possible, provided the contact electrodes evenly sample the underlying nanowire network.

In conclusion, we demonstrated a training-free AI model, the \ac{PANN}, for a natural disaster \ac{EO} change detection task that contained four types of natural disasters: fires, floods, hurricanes and landslides. We benchmarked our model against a state-of-the-art AI model and achieved better results for floods and comparable results in the remaining three categories, without any training. We also showed that the PANN model can generalise to other EO sensors. We found that the \ac{PANN} was able to achieve these results due to its ability to extract meaningful features from the inputs and sort them accordingly in the feature space. Being able to navigate the feature space, therefore, plays a vital role in the model's performance. The training-free nature of the \ac{PANN} makes it an ideal candidate for use on-board satellites, processing \ac{EO} data at the extreme edge.

\section*{Methods}
\label{sec:Methods}

\subsection*{Data}

\begin{figure}
\centering
\includegraphics[width=\linewidth]{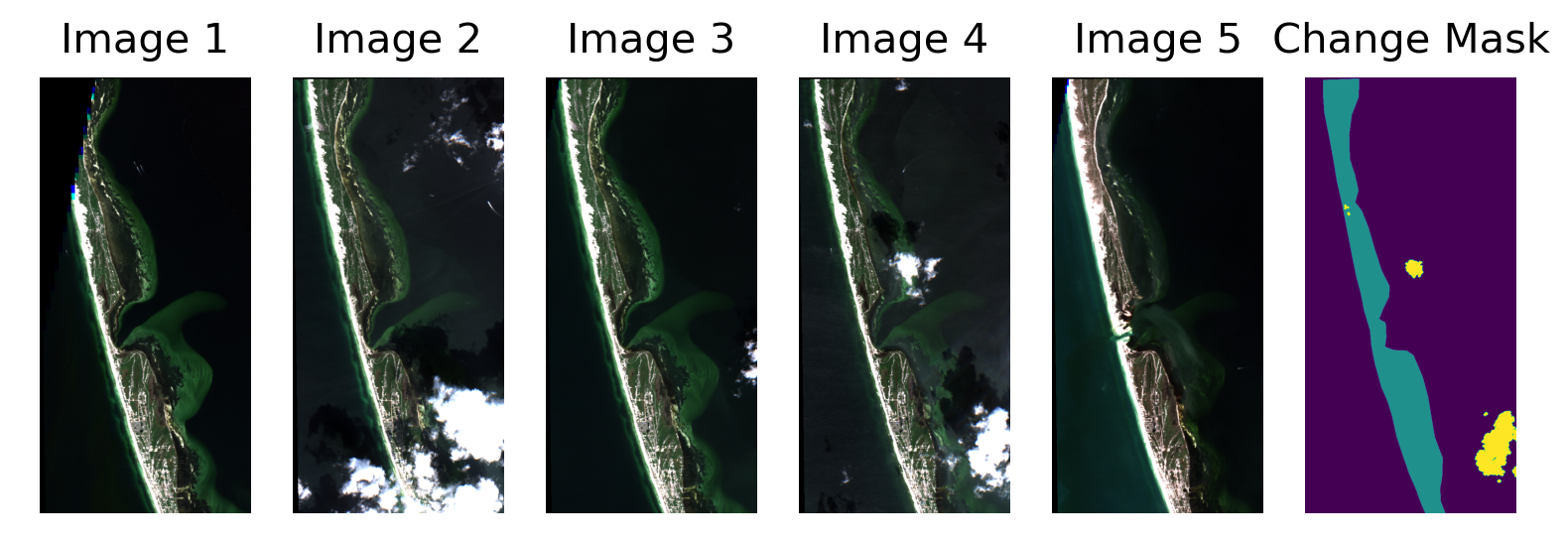}
\caption{An example of a hurricane natural disaster scene, with the first 4 images before the hurricane and image 5 is after. The change mask (last column) shows the affected (aqua) and unaffected (purple) regions along with the clouds (yellow) in the images directly before and after the hurricane. Satellite images are from Sentinel-2 complied by R{\r{u}}{\v{z}}i{\v{c}}ka \textit{et al.} and available in their git\cite{ravaenData}.}
\label{fig:data}
\end{figure}

The primary dataset used in this study is a time series of multi--spectral \ac{EO} of natural disasters released by R{\r{u}}{\v{z}}i{\v{c}}ka \textit{et al.} \cite{ravaen}. The images are level 1C pre--processed multi-spectral images from the Multi-Spectral Imager \cite{drusch2012sentinelMSI} camera on-board the Sentinel-2 Earth observation satellite constellation. Only the 10 highest spatial resolution bands are in the dataset. All bands have a spatial resolution of 10\,m. The bands that are not naturally at 10\,m resolution were interpolated to the higher resolution. 
Table~\ref{tab:bands} shows the specific bands used in this study along with the corresponding index number used to reference the bands.
The dataset covers four categories of natural disasters: fires, floods, hurricanes and landslides. Each category includes five natural disaster events, except for floods which have four, giving a total of 19 natural disaster events. The individual samples were created by tiling each event, with no overlap, into $32 \times 32$ pixel sequence tiles, resulting in a total of 62,655 samples used to evaluate the \ac{PANN} model. For each natural disaster event there are five sequential images; the first four are all before the natural disaster event and the final image is after. Along with each natural disaster event, there is an accompanying change mask that contains three labels: unaffected regions, affected regions, and clouds from the fourth and fifth images. An example of a hurricane disaster event is shown in Fig.~\ref{fig:data}, where damage to the island vegetation can be seen in the fifth image and the change mask image shows the unaffected regions in purple, the affected region in aqua and the clouds in yellow. The change mask in this study is only used to evaluate the performance of the proposed model.

In this study, we created a second dataset to evaluate the performance of the \ac{PANN} from a different sensor, the LandSat-8 Operational Land Imager \cite{knight2014landsatOLI}. This was done by selecting the natural disaster events in the Sentinel-2 dataset that did not have any clouds obscuring the change regions in the change mask. This left 8 natural disaster events: 2 fires, 2 floods and 4 landslides. The Top of Atmosphere images from the LandSat-8 satellite with the 9 highest spatial resolution bands of 30 m were used. To create the time series of images for each natural disaster event, only images with 40\% cloud cover or less were used. Like the Sentinel-2 dataset, each natural disaster event has five sequential images, with the first four before the event and the final image is after. The corresponding change mask from the Sentinel-2 dataset was used after resizing to 30 m resolution using the nearest-neighbour method and adding the clouds in the fourth and fifth images only. Clouds were detected automatically using the Google Earth Engine LandSat simple cloud score algorithm \cite{gorelick2017google}, with a score $\ge 55$\% considered a cloud. As with the Sentinel-2 dataset, individual samples were created by tiling each event, with no overlap, into $16 \times 16$ pixel sequence tiles, resulting in a total of 8,199 samples.

\begin{table}
\centering
\begin{tabular}{|l|l|l|l|l|l|l|l|l|l|l|}
\hline
\multirow{2}{*}{Satellite} & \multicolumn{10}{|c|}{Index (m)} \\
\cline{2-11}
 & 01 & 02 & 03 & 04 & 05 & 06 & 07 & 08 & 09 & 10 \\
\hline
Sentinel-2 & B2 & B3 & B4 & B5 & B6 & B7 & B8 & B8a & B11 & B12 \\
\hline
LandSat-8 & B2 & B3 & B4 & B5 & B6 & B7 & B9 & B10 & B11 & \\
\hline
\end{tabular}
\caption{Sentinel-2 and LandSat-8 bands used in this study and the corresponding index used to reference each band.}
\label{tab:bands}
\end{table}

The data was preprocessed by first taking the log of the RGB pixel values and then scaling the values to the interval [-1, +1] for each band in the data. This process is described by equation~(\ref{eqn:normalisation}):

\begin{equation}
    x^\dagger = \log(x) \qquad , \qquad
    x^\ddagger = 2 \qty( \frac{x^\dagger -\mbox{min}(x^\dagger)}{\mbox{max}(x^\dagger) - \mbox{min}(x^\dagger)} ) -1
\label{eqn:normalisation}
\end{equation}
where the min and max values for each band are fixed and were chosen to be consistent with Ref.~\cite{ravaen} for the Sentinel-2 dataset, to allow a direct comparison with their state-of-the-art variational autoencoder model. A different set of min and max values were manually selected by visual inspection for the LandSat-8 dataset. Missing values were set to a value just above zero of 0.005.

\subsection*{Model Architecture}
 
\begin{figure}
\centering
\begin{subfigure}{0.48\linewidth}
\centering
\includegraphics[width=\linewidth]{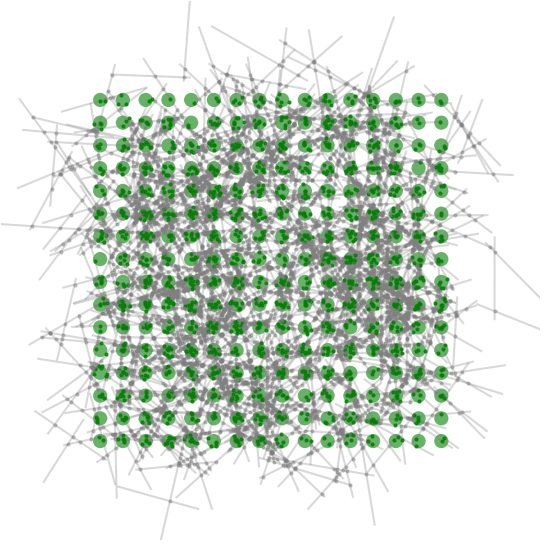}
\caption{}
\label{fig:nwnPANN}
\end{subfigure}
~
\begin{subfigure}{0.48\linewidth}
\centering
\includegraphics[width=\linewidth]{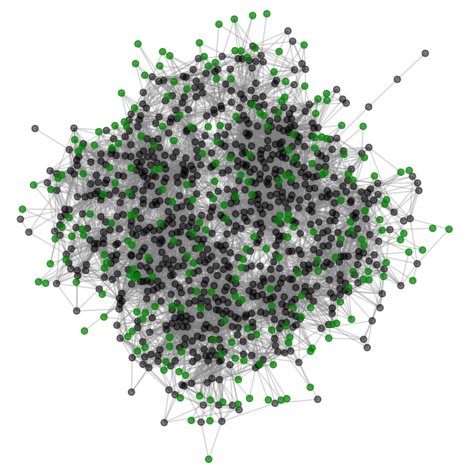}
\caption{}
\label{fig:graphPANN}
\end{subfigure}
\caption{(a) \ac{PANN} model nanowire representation. The nanowires are the grey lines and the electrodes are the large green circles. Nanowire--nanowire junctions are shown as grey dots and nanowire--electrode junctions are shown as dark green dots. (b) graph representation of the \ac{PANN}. Green nodes are the input nodes which correspond to the electrodes. Black nodes correspond to the nanowires, while edges correspond to the nanowire--nanowire and nanowire--electrode junctions.}
\label{fig:networkCreation}
\end{figure}

The PANN is based on a physically--motivated model of a physical neuromorphic network comprised of self--organised nanowires with resistive switching memory (memristive) electrical junctions~\cite{kuncic2020neuromorphic}. This is done by first modelling a nanowire network and then converting it into a graph representation which is used in the \ac{PANN}. This process is shown in Fig.~\ref{fig:networkCreation}. 
In this study, the network for the \ac{PANN} was created by distributing 803 nanowires over a $158 \times 158 \, \mu $m 2D plane with nanowire centres sampled from a generalised normal distribution with a beta value of 5 and with nanowire orientations sampled on $[0,\pi]$. Nanowire lengths were sampled from a Gaussian distribution with an average of $30 \, \mu $m and a standard deviation of $6 \, \mu $m. As nanowire networks are nano-electronic systems, input signals are delivered via electrodes, which were modelled with a diameter $5 \, \mu $m and with evenly spaced placements in a $16\times 16$ grid over the 2D plane with a margin of $15 \, \mu $m, giving a pitch of $8 \, \mu$m between electrodes. Where the nanowires overlap with other nanowires or electrodes, electrical junctions are formed whose internal states evolve in time, as described below. Figure~\ref{fig:nwnPANN} depicts the nanowire network model, with the nanowires displayed as grey lines and nanowire--nanowire junctions as small grey dots. The electrodes are shown as the large green circles and the nanowire--electrode junctions are shown as small dark green dots. 
Figure~\ref{fig:graphPANN} shows the corresponding graphical representation of the network used in the \ac{PANN}. The black nodes correspond to the nanowires and the green nodes correspond to the electrodes, used as input nodes. The edges correspond to the electrical junctions.
The resulting network had a total of 1059 nodes and 12,279 edges. 256 nodes were selected as input nodes and from the remaining 803 nodes, 400 were selected at random as output nodes (note that in a real physical device, electrodes would also serve as output nodes). 

The network in the \ac{PANN} has a heterogeneous, neuromorphic topology and operates like a complex electrical circuit with nonlinear circuit components known as memristors \cite{Loeffler2020,zhu2021information}. As such, inputs are treated as input voltage signals and Kirchhoff's conservation laws are solved at each time step to calculate the node voltage distribution across the network. The edge weights are conductances that evolve in time according to a memristor equation of state:

\begin{align}
    \dv{\lambda}{t} = \begin{cases}
        \qty(|V(t)| -V_{\text{set}})\text{sgn}[V(t)] \; , & |V(t)| > V_{\text{set}} \\
        0 \; , & V_{\text{reset}} < |V(t)| < V_{\text{reset}} \\
        \qty(|V(t)| - V_{\text{reset}})\text{sgn}[\lambda(t)] \; , & |V(t)| < V_{\text{reset}} \\
        0 \; , & |\lambda| \ge \lambda_{\text{max}} \\
    \end{cases}
\label{eqn:filamentState}
\end{align}
with all edges set to an initial state $\lambda(t=0) = 0$.
In this study, the following parameters were used: $V_{\text{reset}} = 5\times10^{-3}\,$V, $V_{\text{set}} = 1\times10^{-2}\,$V and $\lambda_{\text{max}} = 1.5\times10^{-2}\,$V\,s. Full details of the model and its validation against experimental nanowire networks are provided in Refs.~\cite{hochstetter2021avalanches,zhu2021information}. In contrast to artificial neural networks, the dynamic nature of these physical neuromorphic networks means that the weights are not trained, which not only makes them more energy efficient, but also allows all the data to be used for evaluating the model.

\subsection*{Model Setup}

\begin{figure}
\centering
\includegraphics[width=0.95\linewidth]{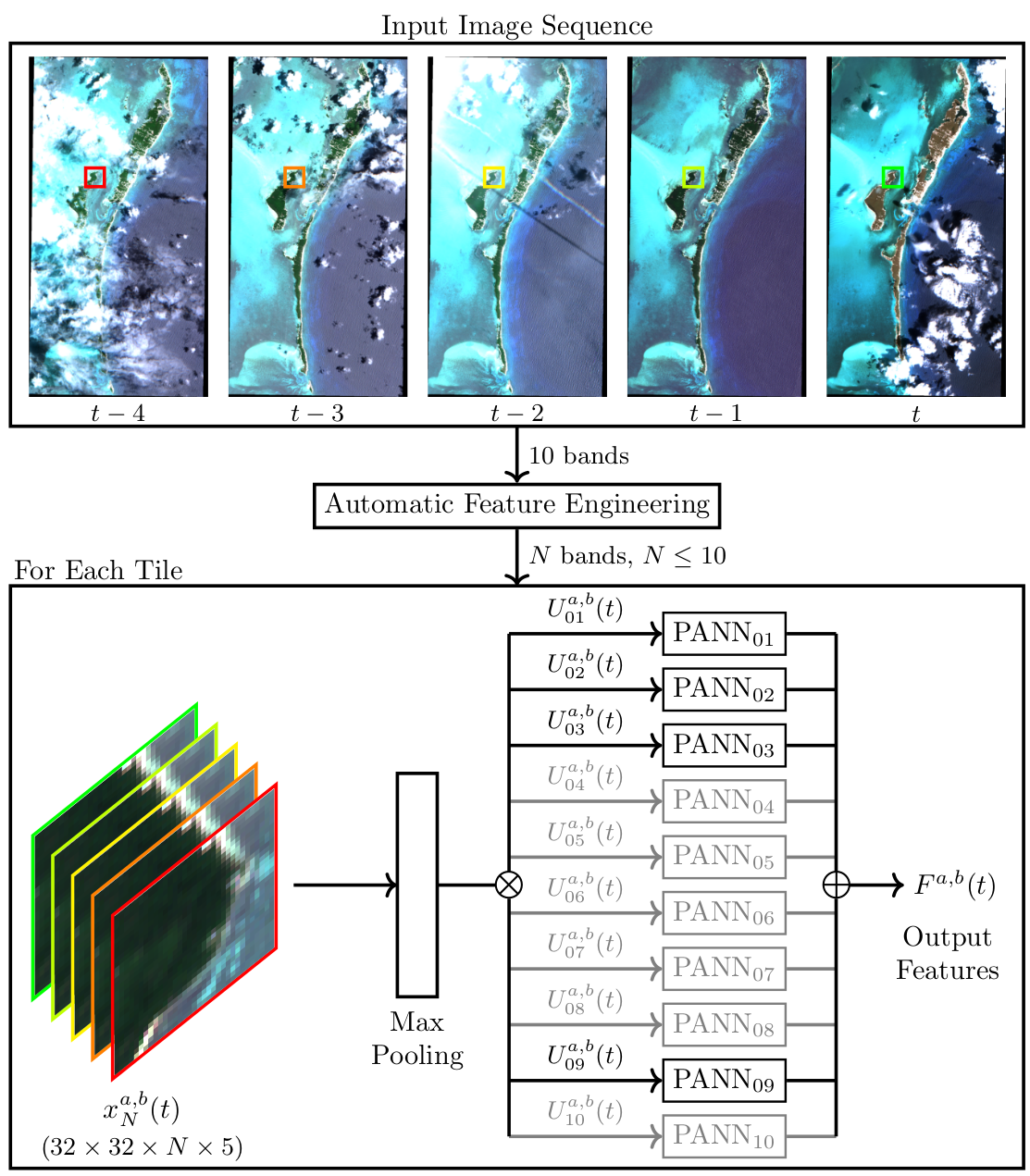}
\caption{Diagram of the \ac{PANN} workflow showing the pipeline from the input images to the output features, $F^{a,b}(t)$ from the networks. First, automatic feature engineering is applied to the input images to determine which bands of the images to use. The images are then broken into a sequence of small $32\times32$ pixel tiles, $x^{a,b} _N(t)$, where $a,b$ is the location of the tile and $N$ is the number of bands. The creation of the tile sequence is shown by the coloured (red, orange, yellow, lime and green) boxes and are then fed into the max pooling layer. The bands are separated, as denoted by $\bigotimes$, with each band being fed into a unique \ac{PANN} model. In this example PANN$_{01-03}$ and PANN$_{09}$ are used. The outputs from all the \acp{PANN} that are used are concatenated together, as denoted by $\bigoplus$, to produce the feature sequence $F^{a,b}(t)$ for tile $x^{a,b} _N(t)$. Satellite images are from Sentinel-2 complied by R{\r{u}}{\v{z}}i{\v{c}}ka \textit{et al.} and available in their git\cite{ravaenData}.}
\label{fig:workflow}
\end{figure}

Figure~\ref{fig:workflow} shows the process of feeding the natural disaster image sequence into the \ac{PANN} model. First an automatic feature engineering step determines which bands to use for that specific natural disaster event. This is motivated by a physical understanding of the event: for example, water exhibits the greatest contrast compared to land at infrared wavelengths. The images are then subdivided into a series of $32\times 32$ pixel sequence tiles, $x^{a,b} _N(t)$, where $a,b$ is the location of the tile and $N \leq 10$ is the number of bands as determined by the feature engineering process. The images are tiled column by column starting at the top left hand corner.
The tile sequence is passed through a max pooling layer that has a pooling size (2,2), a stride of 2 and does not use any padding, hence halving the spatial resolution of the tile sequence to $16\times16$ pixels. The tile sequence is then split into individual bands, creating the input signals $U^{a,b}_m (t)$, where $m$ is the index to each \ac{PANN} for each spectral band as given in Table~\ref{tab:bands}.
As such, there are a total of 10 identical \ac{PANN} networks, each taking inputs for a specific band. The readout values of the \acp{PANN} used for the given tile sequence are concatenated to give the final features, $F^{a,b}(t)$ for that tile location. Hence, not all 10 \acp{PANN} are used for every natural disaster event, as demonstrated in Fig.~\ref{fig:workflow}, where only \ac{PANN} numbers 01-03 and 09 are used and the greyed out \acp{PANN} (PANN 04 -- 08 and 10) are not used for this natural disaster event.

To determine whether a change from a natural disaster occurred at a given tile location, the distance between the images of the tile sequence is measured in the feature space. It is assumed that if there is any change it is in the last image of the sequence, so only the distances between the final image and each of the previous images are compared. This is the same change score method as introduced in Ref.~\cite{ravaen}. Formally, the change score, $S$, is given by
\begin{equation}
    S(x^{a,b}(t)) = \min_{i=1,2,3,4} \left[ \mbox{dist}(F^{a,b}(t-i),F^{a,b}(t)) \right] \qquad ,
\label{eqn:changeScore}
\end{equation}
where $dist$ is an arbitrary distance metric. In this study we focus on three different distance metrics: Euclidean, Cosine and Correlation. Unless otherwise stated, the Correlation distance was used. The Cosine distance is an angle-base metric and is given by
\begin{equation}
    \text{CosineDist}(\mathbf{u},\mathbf{v}) = 1 - \frac{\mathbf{u}\cdot \mathbf{v}}{||\mathbf{u}|| * ||\mathbf{v}||} \qquad .
\label{eqn:cosineDist}
\end{equation}
The Correlation distance is the mean corrected version of the Cosine distance, given by
\begin{equation}
    \text{CorrelationDist}(\mathbf{u},\mathbf{v}) = \text{CosineDist}(\mathbf{u} - \text{mean}(\mathbf{u}),\mathbf{v}- \text{mean}(\mathbf{v})) \qquad .
\label{eqn:corrDist}
\end{equation}
Distances between the tile sequence images are compared directly in the features space, so that it is not necessary to learn a mapping function from the feature space to a target variable. This means it is not necessary to train any part of the model.

To evaluate the quality of the change maps produced by the \ac{PANN} we use the \ac{AUPRC} \cite{davis2006rprecisionRecall}. To calculate the curve, we treated each pixel in the output change maps as a positive or negative example, across all natural disaster events in that category. As such, a separate precision-recall curve was produced for each class of natural disaster and an \ac{AUPRC} value. Areas labelled as clouds in the change mask were ignored when calculating the \ac{AUPRC}. This was done to be consistent with Ref.~\cite{ravaen} and to allow for a more accurate comparison between the \ac{PANN} and RaVAEn models.

\subsection*{Automatic Feature Engineering}

\begin{figure}
\centering
\includegraphics[width=\linewidth]{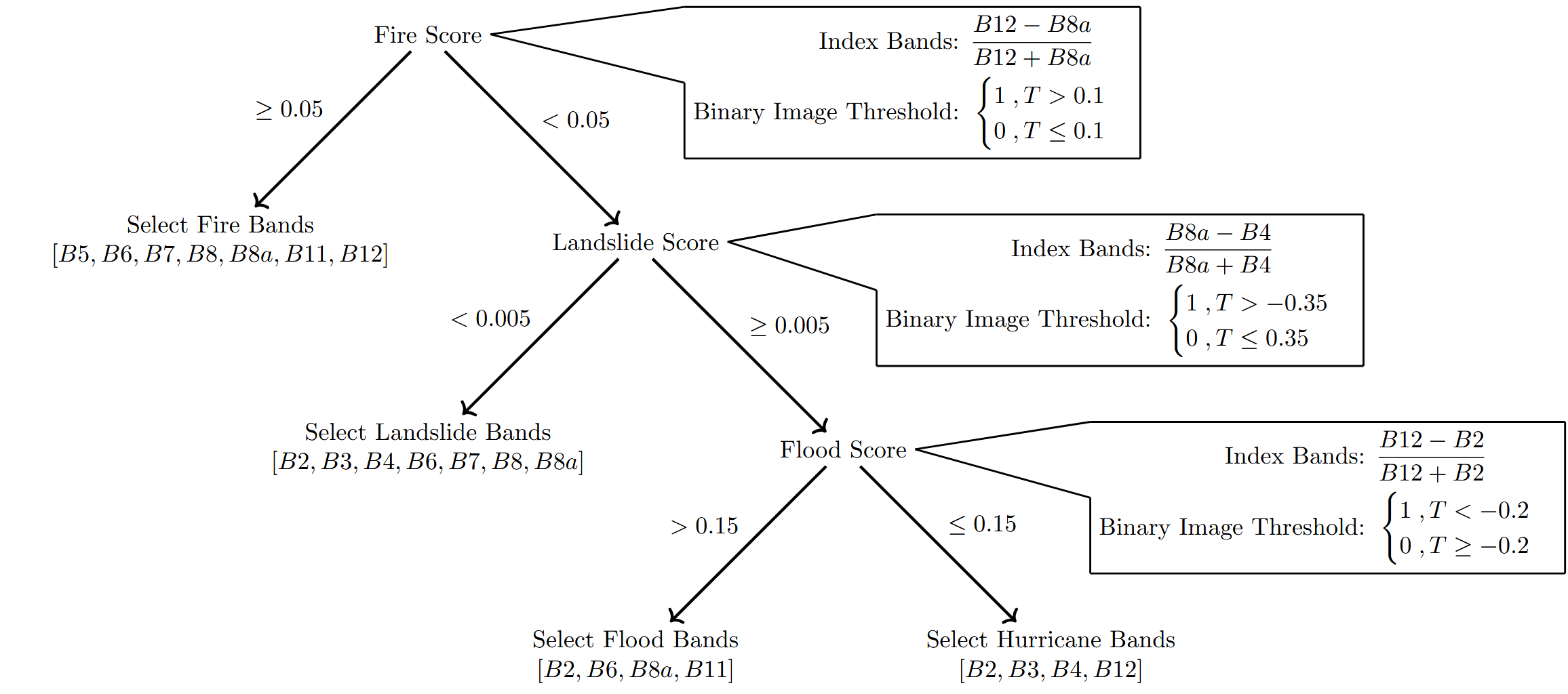}
\caption{Decision tree for the automatic feature engineering step. At each branch a natural disaster score is calculated for the different categories. For each score the bands and threshold values used to create the index image are given. The bands used for each natural disaster type once selected are also shown.}
\label{fig:bandSelector}
\end{figure}

The automatic feature engineering process is performed once for each natural disaster scene and is performed first, before the normalisation and tiling stages. The feature engineering process leverages common indices such as the normalised burn ratio index \cite{garcia1991mapping} and the normalised difference vegetation index \cite{ndviCommentary} to calculate natural disaster class scores. 
Using these scores, a binary decision tree is followed, as shown in Fig.~\ref{fig:bandSelector}, to determine what class of natural disaster is in the scene.
The score for each natural disaster class is calculated by creating an index image for the images directly before and after the natural disaster event using equation~(\ref{eqn:index}), where $BX$ and $BY$ are predetermined bands depending on what type of natural disaster is being assessed. The images are converted to a binary image using a predefined threshold value and a difference image is created by a pixel--wise comparison. The final score is the percentage of pixels that record a change. Once a natural disaster type has been selected, the feature engineering process then selects a subset of bands to be used in the rest of the model depending on the natural disaster class. The specific values and bands used for the natural disaster class score are shown in Fig.~\ref{fig:bandSelector}. See Supplementary Fig.~S4 online for the specific values and bands used for the LandSat-8 data.

\begin{equation}
    \frac{BX - BY}{BX + BY}
\label{eqn:index}
\end{equation}

We note that more sophisticated index images \cite{filipponi2018bais2, muhsoni2018comparison} could be used to categorise the natural disasters and potentially further improve results. Furthermore, similar feature engineering processes could be used to tune the model to detect changes from specific natural disasters or even clouds. The focus here is to show that the simplest automatic feature engineering process, one that does not require training, can be used to easily incorporate domain knowledge and improve the performance of the model. 

\section*{Data Availability}

We are releasing the full code alongside this paper at \url{https://github.com/ssmi9157/ChangeDetectionPANN}. The Sentinel-2 dataset analysed in this study is available in the RaVAEn repository \url{https://github.com/spaceml-org/RaVAEn} and the LandSat-8 dataset is available at \url{https://zenodo.org/records/16916355}.

\bibliography{sample}

\begin{thebibliography}{10}
\urlstyle{rm}
\expandafter\ifx\csname url\endcsname\relax
  \def\url#1{\texttt{#1}}\fi
\expandafter\ifx\csname urlprefix\endcsname\relax\def\urlprefix{URL }\fi
\expandafter\ifx\csname doiprefix\endcsname\relax\def\doiprefix{DOI: }\fi
\providecommand{\bibinfo}[2]{#2}
\providecommand{\eprint}[2][]{\url{#2}}

\bibitem{vali2020deep}
\bibinfo{author}{Vali, A.}, \bibinfo{author}{Comai, S.} \&
  \bibinfo{author}{Matteucci, M.}
\newblock \bibinfo{journal}{\bibinfo{title}{Deep learning for land use and land
  cover classification based on hyperspectral and multispectral earth
  observation data: A review}}.
\newblock {\emph{\JournalTitle{Remote Sensing}}} \textbf{\bibinfo{volume}{12}},
  \bibinfo{pages}{2495} (\bibinfo{year}{2020}).

\bibitem{maguluri2024sustainable}
\bibinfo{author}{Maguluri, L.~P.} \emph{et~al.}
\newblock \bibinfo{journal}{\bibinfo{title}{Sustainable agriculture and climate
  change: A deep learning approach to remote sensing for food security
  monitoring}}.
\newblock {\emph{\JournalTitle{Remote Sensing in Earth Systems Sciences}}}
  \bibinfo{pages}{1--13} (\bibinfo{year}{2024}).

\bibitem{ruuvzivcka2023semantic}
\bibinfo{author}{R{\r{u}}{\v{z}}i{\v{c}}ka, V.} \emph{et~al.}
\newblock \bibinfo{journal}{\bibinfo{title}{Semantic segmentation of methane
  plumes with hyperspectral machine learning models}}.
\newblock {\emph{\JournalTitle{Scientific Reports}}}
  \textbf{\bibinfo{volume}{13}}, \bibinfo{pages}{19999} (\bibinfo{year}{2023}).

\bibitem{ferreira2020monitoring}
\bibinfo{author}{Ferreira, B.}, \bibinfo{author}{Iten, M.} \&
  \bibinfo{author}{Silva, R.~G.}
\newblock \bibinfo{journal}{\bibinfo{title}{Monitoring sustainable development
  by means of earth observation data and machine learning: A review}}.
\newblock {\emph{\JournalTitle{Environmental Sciences Europe}}}
  \textbf{\bibinfo{volume}{32}}, \bibinfo{pages}{1--17} (\bibinfo{year}{2020}).

\bibitem{parelius2023review}
\bibinfo{author}{Parelius, E.~J.}
\newblock \bibinfo{journal}{\bibinfo{title}{A review of deep-learning methods
  for change detection in multispectral remote sensing images}}.
\newblock {\emph{\JournalTitle{Remote Sensing}}} \textbf{\bibinfo{volume}{15}},
  \bibinfo{pages}{2092} (\bibinfo{year}{2023}).

\bibitem{diakogiannis2021looking}
\bibinfo{author}{Diakogiannis, F.~I.}, \bibinfo{author}{Waldner, F.} \&
  \bibinfo{author}{Caccetta, P.}
\newblock \bibinfo{journal}{\bibinfo{title}{Looking for change? roll the dice
  and demand attention}}.
\newblock {\emph{\JournalTitle{Remote Sensing}}} \textbf{\bibinfo{volume}{13}},
  \bibinfo{pages}{3707} (\bibinfo{year}{2021}).

\bibitem{chen2020spatial}
\bibinfo{author}{Chen, H.} \& \bibinfo{author}{Shi, Z.}
\newblock \bibinfo{journal}{\bibinfo{title}{A spatial-temporal attention-based
  method and a new dataset for remote sensing image change detection}}.
\newblock {\emph{\JournalTitle{Remote sensing}}} \textbf{\bibinfo{volume}{12}},
  \bibinfo{pages}{1662} (\bibinfo{year}{2020}).

\bibitem{bello2014satellite}
\bibinfo{author}{Bello, O.~M.} \& \bibinfo{author}{Aina, Y.~A.}
\newblock \bibinfo{journal}{\bibinfo{title}{Satellite remote sensing as a tool
  in disaster management and sustainable development: towards a synergistic
  approach}}.
\newblock {\emph{\JournalTitle{Procedia-Social and Behavioral Sciences}}}
  \textbf{\bibinfo{volume}{120}}, \bibinfo{pages}{365--373}
  (\bibinfo{year}{2014}).

\bibitem{huyck2014remote}
\bibinfo{author}{Huyck, C.}, \bibinfo{author}{Verrucci, E.} \&
  \bibinfo{author}{Bevington, J.}
\newblock \bibinfo{title}{Remote sensing for disaster response: A rapid,
  image-based perspective}.
\newblock In \emph{\bibinfo{booktitle}{Earthquake hazard, risk and disasters}},
  \bibinfo{pages}{1--24} (\bibinfo{publisher}{Elsevier}, \bibinfo{year}{2014}).

\bibitem{le2020space}
\bibinfo{author}{Le~Cozannet, G.} \emph{et~al.}
\newblock \bibinfo{journal}{\bibinfo{title}{Space-based earth observations for
  disaster risk management}}.
\newblock {\emph{\JournalTitle{Surveys in geophysics}}}
  \textbf{\bibinfo{volume}{41}}, \bibinfo{pages}{1209--1235}
  (\bibinfo{year}{2020}).

\bibitem{karapetyan2015satellite}
\bibinfo{author}{Karapetyan, D.}, \bibinfo{author}{Minic, S.~M.},
  \bibinfo{author}{Malladi, K.~T.} \& \bibinfo{author}{Punnen, A.~P.}
\newblock \bibinfo{journal}{\bibinfo{title}{Satellite downlink scheduling
  problem: A case study}}.
\newblock {\emph{\JournalTitle{Omega}}} \textbf{\bibinfo{volume}{53}},
  \bibinfo{pages}{115--123} (\bibinfo{year}{2015}).

\bibitem{selva2012survey}
\bibinfo{author}{Selva, D.} \& \bibinfo{author}{Krejci, D.}
\newblock \bibinfo{journal}{\bibinfo{title}{A survey and assessment of the
  capabilities of cubesats for earth observation}}.
\newblock {\emph{\JournalTitle{Acta Astronautica}}}
  \textbf{\bibinfo{volume}{74}}, \bibinfo{pages}{50--68}
  (\bibinfo{year}{2012}).

\bibitem{verbesselt2010detecting}
\bibinfo{author}{Verbesselt, J.}, \bibinfo{author}{Hyndman, R.},
  \bibinfo{author}{Newnham, G.} \& \bibinfo{author}{Culvenor, D.}
\newblock \bibinfo{journal}{\bibinfo{title}{Detecting trend and seasonal
  changes in satellite image time series}}.
\newblock {\emph{\JournalTitle{Remote sensing of Environment}}}
  \textbf{\bibinfo{volume}{114}}, \bibinfo{pages}{106--115}
  (\bibinfo{year}{2010}).

\bibitem{rodriguez2024monitoring}
\bibinfo{author}{Rodriguez, P.~S.}, \bibinfo{author}{Schwantes, A.~M.},
  \bibinfo{author}{Gonzalez, A.} \& \bibinfo{author}{Fortin, M.-J.}
\newblock \bibinfo{journal}{\bibinfo{title}{Monitoring changes in the enhanced
  vegetation index to inform the management of forests}}.
\newblock {\emph{\JournalTitle{Remote Sensing}}} \textbf{\bibinfo{volume}{16}},
  \bibinfo{pages}{2919} (\bibinfo{year}{2024}).

\bibitem{normalizedBurnRatio}
\bibinfo{author}{Alcaras, E.}, \bibinfo{author}{Costantino, D.},
  \bibinfo{author}{Guastaferro, F.}, \bibinfo{author}{Parente, C.} \&
  \bibinfo{author}{Pepe, M.}
\newblock \bibinfo{journal}{\bibinfo{title}{Normalized burn ratio plus (nbr+):
  a new index for sentinel-2 imagery}}.
\newblock {\emph{\JournalTitle{Remote Sensing}}} \textbf{\bibinfo{volume}{14}},
  \bibinfo{pages}{1727} (\bibinfo{year}{2022}).

\bibitem{zheng2018new}
\bibinfo{author}{Zheng, Q.}, \bibinfo{author}{Huang, W.}, \bibinfo{author}{Cui,
  X.}, \bibinfo{author}{Shi, Y.} \& \bibinfo{author}{Liu, L.}
\newblock \bibinfo{journal}{\bibinfo{title}{New spectral index for detecting
  wheat yellow rust using sentinel-2 multispectral imagery}}.
\newblock {\emph{\JournalTitle{Sensors}}} \textbf{\bibinfo{volume}{18}},
  \bibinfo{pages}{868} (\bibinfo{year}{2018}).

\bibitem{farhadi2025introducing}
\bibinfo{author}{Farhadi, H.}, \bibinfo{author}{Ebadi, H.},
  \bibinfo{author}{Kiani, A.} \& \bibinfo{author}{Asgary, A.}
\newblock \bibinfo{journal}{\bibinfo{title}{Introducing a new index for flood
  mapping using sentinel-2 imagery (sfmi)}}.
\newblock {\emph{\JournalTitle{Computers \& Geosciences}}}
  \textbf{\bibinfo{volume}{194}}, \bibinfo{pages}{105742}
  (\bibinfo{year}{2025}).

\bibitem{pettorelli2013ndviBook}
\bibinfo{author}{Pettorelli, N.}
\newblock \emph{\bibinfo{title}{The normalized difference vegetation index}}
  (\bibinfo{publisher}{Oxford University Press, USA}, \bibinfo{year}{2013}).

\bibitem{jiang2008development}
\bibinfo{author}{Jiang, Z.}, \bibinfo{author}{Huete, A.~R.},
  \bibinfo{author}{Didan, K.} \& \bibinfo{author}{Miura, T.}
\newblock \bibinfo{journal}{\bibinfo{title}{Development of a two-band enhanced
  vegetation index without a blue band}}.
\newblock {\emph{\JournalTitle{Remote sensing of Environment}}}
  \textbf{\bibinfo{volume}{112}}, \bibinfo{pages}{3833--3845}
  (\bibinfo{year}{2008}).

\bibitem{nemni2020fully}
\bibinfo{author}{Nemni, E.}, \bibinfo{author}{Bullock, J.},
  \bibinfo{author}{Belabbes, S.} \& \bibinfo{author}{Bromley, L.}
\newblock \bibinfo{journal}{\bibinfo{title}{Fully convolutional neural network
  for rapid flood segmentation in synthetic aperture radar imagery}}.
\newblock {\emph{\JournalTitle{Remote Sensing}}} \textbf{\bibinfo{volume}{12}},
  \bibinfo{pages}{2532} (\bibinfo{year}{2020}).

\bibitem{rudner2019multi3net}
\bibinfo{author}{Rudner, T.~G.} \emph{et~al.}
\newblock \bibinfo{title}{Multi3net: segmenting flooded buildings via fusion of
  multiresolution, multisensor, and multitemporal satellite imagery}.
\newblock In \emph{\bibinfo{booktitle}{Proceedings of the AAAI Conference on
  Artificial Intelligence}}, vol.~\bibinfo{volume}{33},
  \bibinfo{pages}{702--709} (\bibinfo{year}{2019}).

\bibitem{qing2022operational}
\bibinfo{author}{Qing, Y.} \emph{et~al.}
\newblock \bibinfo{journal}{\bibinfo{title}{Operational earthquake-induced
  building damage assessment using cnn-based direct remote sensing change
  detection on superpixel level}}.
\newblock {\emph{\JournalTitle{International Journal of Applied Earth
  Observation and Geoinformation}}} \textbf{\bibinfo{volume}{112}},
  \bibinfo{pages}{102899} (\bibinfo{year}{2022}).

\bibitem{del2021board}
\bibinfo{author}{Del~Rosso, M.~P.}, \bibinfo{author}{Sebastianelli, A.},
  \bibinfo{author}{Spiller, D.}, \bibinfo{author}{Mathieu, P.~P.} \&
  \bibinfo{author}{Ullo, S.~L.}
\newblock \bibinfo{journal}{\bibinfo{title}{On-board volcanic eruption
  detection through cnns and satellite multispectral imagery}}.
\newblock {\emph{\JournalTitle{Remote Sensing}}} \textbf{\bibinfo{volume}{13}},
  \bibinfo{pages}{3479} (\bibinfo{year}{2021}).

\bibitem{chintalapati2024opportunitiesOnboardAI}
\bibinfo{author}{Chintalapati, B.} \emph{et~al.}
\newblock \bibinfo{journal}{\bibinfo{title}{Opportunities and challenges of
  on-board ai-based image recognition for small satellite earth observation
  missions}}.
\newblock {\emph{\JournalTitle{Advances in Space Research}}}
  (\bibinfo{year}{2024}).

\bibitem{yu2022coca}
\bibinfo{author}{Yu, J.} \emph{et~al.}
\newblock \bibinfo{journal}{\bibinfo{title}{Coca: Contrastive captioners are
  image-text foundation models}}.
\newblock {\emph{\JournalTitle{arXiv preprint arXiv:2205.01917}}}
  (\bibinfo{year}{2022}).

\bibitem{chen2023symbolic}
\bibinfo{author}{Chen, X.} \emph{et~al.}
\newblock \bibinfo{journal}{\bibinfo{title}{Symbolic discovery of optimization
  algorithms}}.
\newblock {\emph{\JournalTitle{Advances in neural information processing
  systems}}} \textbf{\bibinfo{volume}{36}}, \bibinfo{pages}{49205--49233}
  (\bibinfo{year}{2023}).

\bibitem{abid2021ucl}
\bibinfo{author}{Abid, N.} \emph{et~al.}
\newblock \bibinfo{journal}{\bibinfo{title}{Ucl: Unsupervised curriculum
  learning for water body classification from remote sensing imagery}}.
\newblock {\emph{\JournalTitle{International Journal of Applied Earth
  Observation and Geoinformation}}} \textbf{\bibinfo{volume}{105}},
  \bibinfo{pages}{102568} (\bibinfo{year}{2021}).

\bibitem{ravaen}
\bibinfo{author}{R{\r{u}}{\v{z}}i{\v{c}}ka, V.} \emph{et~al.}
\newblock \bibinfo{journal}{\bibinfo{title}{Rav{\ae}n: unsupervised change
  detection of extreme events using ml on-board satellites}}.
\newblock {\emph{\JournalTitle{Scientfic reports}}}
  \textbf{\bibinfo{volume}{12}}, \bibinfo{pages}{16939} (\bibinfo{year}{2022}).

\bibitem{giuffrida2021varphi}
\bibinfo{author}{Giuffrida, G.} \emph{et~al.}
\newblock \bibinfo{journal}{\bibinfo{title}{The $\phi$-sat-1 mission: The first
  on-board deep neural network demonstrator for satellite earth observation}}.
\newblock {\emph{\JournalTitle{IEEE Transactions on Geoscience and Remote
  Sensing}}} \textbf{\bibinfo{volume}{60}}, \bibinfo{pages}{1--14}
  (\bibinfo{year}{2021}).

\bibitem{labreche2022ops}
\bibinfo{author}{Labr{\`e}che, G.} \emph{et~al.}
\newblock \bibinfo{title}{Ops-sat spacecraft autonomy with tensorflow lite,
  unsupervised learning, and online machine learning}.
\newblock In \emph{\bibinfo{booktitle}{2022 IEEE Aerospace Conference (AERO)}},
  \bibinfo{pages}{1--17} (\bibinfo{organization}{IEEE}, \bibinfo{year}{2022}).

\bibitem{koh2021wilds}
\bibinfo{author}{Koh, P.~W.} \emph{et~al.}
\newblock \bibinfo{title}{Wilds: A benchmark of in-the-wild distribution
  shifts}.
\newblock In \emph{\bibinfo{booktitle}{International conference on machine
  learning}}, \bibinfo{pages}{5637--5664} (\bibinfo{organization}{PMLR},
  \bibinfo{year}{2021}).

\bibitem{derksen2021fewShotChallenge}
\bibinfo{author}{Derksen, D.} \emph{et~al.}
\newblock \bibinfo{title}{Few-shot image classification challenge on-board}.
\newblock In \emph{\bibinfo{booktitle}{Workshop-data centric AI, neurIPS}}
  (\bibinfo{year}{2021}).

\bibitem{mateo2023inorbitTraining}
\bibinfo{author}{Mateo-Garc{\'\i}a, G.} \emph{et~al.}
\newblock \bibinfo{journal}{\bibinfo{title}{In-orbit demonstration of a
  re-trainable machine learning payload for processing optical imagery}}.
\newblock {\emph{\JournalTitle{Scientific Reports}}}
  \textbf{\bibinfo{volume}{13}}, \bibinfo{pages}{10391} (\bibinfo{year}{2023}).

\bibitem{ruuvzivcka2023fast}
\bibinfo{author}{R{\r{u}}{\v{z}}i{\v{c}}ka, V.} \emph{et~al.}
\newblock \bibinfo{title}{Fast model inference and training on-board of
  satellites}.
\newblock In \emph{\bibinfo{booktitle}{IGARSS 2023-2023 IEEE International
  Geoscience and Remote Sensing Symposium}}, \bibinfo{pages}{2002--2005}
  (\bibinfo{organization}{IEEE}, \bibinfo{year}{2023}).

\bibitem{christensen20222022}
\bibinfo{author}{Christensen, D.~V.} \emph{et~al.}
\newblock \bibinfo{journal}{\bibinfo{title}{2022 roadmap on neuromorphic
  computing and engineering}}.
\newblock {\emph{\JournalTitle{Neuromorphic Computing and Engineering}}}
  \textbf{\bibinfo{volume}{2}}, \bibinfo{pages}{022501} (\bibinfo{year}{2022}).

\bibitem{song2023recent}
\bibinfo{author}{Song, M.-K.} \emph{et~al.}
\newblock \bibinfo{journal}{\bibinfo{title}{Recent advances and future
  prospects for memristive materials, devices, and systems}}.
\newblock {\emph{\JournalTitle{ACS nano}}} \textbf{\bibinfo{volume}{17}},
  \bibinfo{pages}{11994--12039} (\bibinfo{year}{2023}).

\bibitem{Kuncic-Nakayama_2021}
\bibinfo{author}{Kuncic, Z.} \& \bibinfo{author}{Nakayama, T.}
\newblock \bibinfo{journal}{\bibinfo{title}{Neuromorphic nanowire networks:
  Principles, progress and future prospects for neuro-inspired information
  processing}}.
\newblock {\emph{\JournalTitle{Advances in Physics: X}}}
  \textbf{\bibinfo{volume}{6}}, \bibinfo{pages}{1894234}
  (\bibinfo{year}{2021}).

\bibitem{karniadakis2021physics}
\bibinfo{author}{Karniadakis, G.~E.} \emph{et~al.}
\newblock \bibinfo{journal}{\bibinfo{title}{Physics-informed machine
  learning}}.
\newblock {\emph{\JournalTitle{Nature Reviews Physics}}}
  \textbf{\bibinfo{volume}{3}}, \bibinfo{pages}{422--440}
  (\bibinfo{year}{2021}).

\bibitem{cuomo2022scientific}
\bibinfo{author}{Cuomo, S.} \emph{et~al.}
\newblock \bibinfo{journal}{\bibinfo{title}{Scientific machine learning through
  physics--informed neural networks: Where we are and what’s next}}.
\newblock {\emph{\JournalTitle{Journal of Scientific Computing}}}
  \textbf{\bibinfo{volume}{92}}, \bibinfo{pages}{88} (\bibinfo{year}{2022}).

\bibitem{jin2021nsfnets}
\bibinfo{author}{Jin, X.}, \bibinfo{author}{Cai, S.}, \bibinfo{author}{Li, H.}
  \& \bibinfo{author}{Karniadakis, G.~E.}
\newblock \bibinfo{journal}{\bibinfo{title}{Nsfnets (navier-stokes flow nets):
  Physics-informed neural networks for the incompressible navier-stokes
  equations}}.
\newblock {\emph{\JournalTitle{Journal of Computational Physics}}}
  \textbf{\bibinfo{volume}{426}}, \bibinfo{pages}{109951}
  (\bibinfo{year}{2021}).

\bibitem{Zhu2023online}
\bibinfo{author}{Zhu, R.} \emph{et~al.}
\newblock \bibinfo{journal}{\bibinfo{title}{Online dynamical learning and
  sequence memory with neuromorphic nanowire networks}}.
\newblock {\emph{\JournalTitle{Nature Communications}}}
  \textbf{\bibinfo{volume}{14}}, \bibinfo{pages}{6697} (\bibinfo{year}{2023}).

\bibitem{Lilak2021}
\bibinfo{author}{Lilak, S.} \emph{et~al.}
\newblock \bibinfo{journal}{\bibinfo{title}{Spoken digit classification by
  in-materio reservoir computing with neuromorphic atomic switch networks}}.
\newblock {\emph{\JournalTitle{Frontiers in Nanotechnology}}}
  \textbf{\bibinfo{volume}{3}}, \doiprefix\url{10.3389/fnano.2021.675792}
  (\bibinfo{year}{2021}).

\bibitem{Kotooka2024}
\bibinfo{author}{Kotooka, T.} \emph{et~al.}
\newblock \bibinfo{journal}{\bibinfo{title}{Thermally {Stable} {Ag2Se}
  {Nanowire} {Network} as an {Effective} {In}-{Materio} {Physical} {Reservoir}
  {Computing} {Device}}}.
\newblock {\emph{\JournalTitle{Advanced Electronic Materials}}}
  \textbf{\bibinfo{volume}{10}}, \bibinfo{pages}{2400443},
  \doiprefix\url{10.1002/aelm.202400443} (\bibinfo{year}{2024}).

\bibitem{Loeffler2023neuromorphic}
\bibinfo{author}{Loeffler, A.} \emph{et~al.}
\newblock \bibinfo{journal}{\bibinfo{title}{Neuromorphic learning, working
  memory, and metaplasticity in nanowire networks}}.
\newblock {\emph{\JournalTitle{Science Advances}}}
  \textbf{\bibinfo{volume}{9}}, \bibinfo{pages}{eadg3289}
  (\bibinfo{year}{2023}).

\bibitem{Sillin2013theoretical}
\bibinfo{author}{Sillin, H.~O.} \emph{et~al.}
\newblock \bibinfo{journal}{\bibinfo{title}{A theoretical and experimental
  study of neuromorphic atomic switch networks for reservoir computing}}.
\newblock {\emph{\JournalTitle{Nanotechnology}}} \textbf{\bibinfo{volume}{24}},
  \bibinfo{pages}{384004} (\bibinfo{year}{2013}).

\bibitem{Stieg2014_self-org}
\bibinfo{author}{Stieg, A.~Z.} \emph{et~al.}
\newblock \bibinfo{title}{Self-organization and {Emergence} of {Dynamical}
  {Structures} in {Neuromorphic} {Atomic} {Switch} {Networks}}.
\newblock In \bibinfo{editor}{Adamatzky, A.} \& \bibinfo{editor}{Chua, L.}
  (eds.) \emph{\bibinfo{booktitle}{Memristor {Networks}}},
  \bibinfo{pages}{173--209}, \doiprefix\url{10.1007/978-3-319-02630-5_10}
  (\bibinfo{publisher}{Springer International Publishing},
  \bibinfo{year}{2014}).

\bibitem{Michieletti2025soc}
\bibinfo{author}{Michieletti, F.}, \bibinfo{author}{Pilati, D.},
  \bibinfo{author}{Milano, G.} \& \bibinfo{author}{Ricciardi, C.}
\newblock \bibinfo{journal}{\bibinfo{title}{Self-organized criticality in
  neuromorphic nanowire networks with tunable and local dynamics}}.
\newblock {\emph{\JournalTitle{Advanced Functional Materials}}}
  \bibinfo{pages}{2423903} (\bibinfo{year}{2025}).

\bibitem{Milano2022materia}
\bibinfo{author}{Milano, G.} \emph{et~al.}
\newblock \bibinfo{journal}{\bibinfo{title}{In materia reservoir computing with
  a fully memristive architecture based on self-organizing nanowire networks}}.
\newblock {\emph{\JournalTitle{Nature materials}}}
  \textbf{\bibinfo{volume}{21}}, \bibinfo{pages}{195--202}
  (\bibinfo{year}{2022}).

\bibitem{kuncic2020neuromorphic}
\bibinfo{author}{Kuncic, Z.} \emph{et~al.}
\newblock \bibinfo{title}{Neuromorphic information processing with nanowire
  networks}.
\newblock In \emph{\bibinfo{booktitle}{2020 IEEE International Symposium on
  Circuits and Systems (ISCAS)}}, \bibinfo{pages}{1--5}
  (\bibinfo{organization}{IEEE}, \bibinfo{year}{2020}).

\bibitem{Fu2020}
\bibinfo{author}{Fu, K.} \emph{et~al.}
\newblock \bibinfo{title}{Reservoir computing with neuromemristive nanowire
  networks}.
\newblock In \emph{\bibinfo{booktitle}{2020 International Joint Conference on
  Neural Networks (IJCNN)}}, \bibinfo{pages}{1--8},
  \doiprefix\url{10.1109/IJCNN48605.2020.9207727} (\bibinfo{year}{2020}).

\bibitem{hochstetter2021avalanches}
\bibinfo{author}{Hochstetter, J.} \emph{et~al.}
\newblock \bibinfo{journal}{\bibinfo{title}{Avalanches and edge-of-chaos
  learning in neuromorphic nanowire networks}}.
\newblock {\emph{\JournalTitle{Nature Communications}}}
  \textbf{\bibinfo{volume}{12}}, \bibinfo{pages}{4008} (\bibinfo{year}{2021}).

\bibitem{zhu2020harnessing}
\bibinfo{author}{Zhu, R.} \emph{et~al.}
\newblock \bibinfo{title}{Harnessing adaptive dynamics in neuro-memristive
  nanowire networks for transfer learning}.
\newblock In \emph{\bibinfo{booktitle}{2020 International Conference on
  Rebooting Computing (ICRC)}}, \bibinfo{pages}{102--106}
  (\bibinfo{organization}{IEEE}, \bibinfo{year}{2020}).

\bibitem{zhu2021information}
\bibinfo{author}{Zhu, R.} \emph{et~al.}
\newblock \bibinfo{journal}{\bibinfo{title}{Information dynamics in
  neuromorphic nanowire networks}}.
\newblock {\emph{\JournalTitle{Scientific Reports}}}
  \textbf{\bibinfo{volume}{11}}, \bibinfo{pages}{13047} (\bibinfo{year}{2021}).

\bibitem{Loeffler2021}
\bibinfo{author}{Loeffler, A.} \emph{et~al.}
\newblock \bibinfo{journal}{\bibinfo{title}{Modularity and multitasking in
  neuro-memristive reservoir networks}}.
\newblock {\emph{\JournalTitle{Neuromorphic Computing and Engineering}}}
  \textbf{\bibinfo{volume}{1}}, \bibinfo{pages}{014003},
  \doiprefix\url{10.1088/2634-4386/ac156f} (\bibinfo{year}{2021}).

\bibitem{Zhu2023_L2L}
\bibinfo{author}{Zhu, R.}, \bibinfo{author}{Eshraghian, J.} \&
  \bibinfo{author}{Kuncic, Z.}
\newblock \bibinfo{title}{Memristive reservoirs learn to learn}.
\newblock In \emph{\bibinfo{booktitle}{Proceedings of the 2023 International
  Conference on Neuromorphic Systems}}, ICONS '23,
  \doiprefix\url{10.1145/3589737.3605989} (\bibinfo{publisher}{Association for
  Computing Machinery}, \bibinfo{address}{New York, NY, USA},
  \bibinfo{year}{2023}).

\bibitem{Baccetti2024}
\bibinfo{author}{Baccetti, V.}, \bibinfo{author}{Zhu, R.},
  \bibinfo{author}{Kuncic, Z.} \& \bibinfo{author}{Caravelli, F.}
\newblock \bibinfo{journal}{\bibinfo{title}{Ergodicity, lack thereof, and the
  performance of reservoir computing with memristive networks}}.
\newblock {\emph{\JournalTitle{Nano Express}}} \textbf{\bibinfo{volume}{5}},
  \bibinfo{pages}{015021}, \doiprefix\url{10.1088/2632-959X/ad2999}
  (\bibinfo{year}{2024}).

\bibitem{ravaenData}
\bibinfo{author}{R{\r{u}}{\v{z}}i{\v{c}}ka, V.} \emph{et~al.}
\newblock \bibinfo{title}{Ravaen}.
\newblock \bibinfo{howpublished}{\emph{GitHub}
  \url{https://github.com/spaceml-org/RaVAEn}} (\bibinfo{year}{2022}).
\newblock \bibinfo{note}{Accessed: May 1, 2024}.

\bibitem{kingma2013VAE}
\bibinfo{author}{Kingma, D.~P.} \& \bibinfo{author}{Welling, M.}
\newblock \bibinfo{title}{Auto-encoding variational bayes}.
\newblock In \emph{\bibinfo{booktitle}{International Conference on Learning
  Representations}} (\bibinfo{year}{2014}).

\bibitem{mcinnes2018umap}
\bibinfo{author}{McInnes, L.}, \bibinfo{author}{Healy, J.} \&
  \bibinfo{author}{Melville, J.}
\newblock \bibinfo{journal}{\bibinfo{title}{Umap: Uniform manifold
  approximation and projection for dimension reduction}}.
\newblock {\emph{\JournalTitle{arXiv preprint arXiv:1802.03426}}}
  (\bibinfo{year}{2018}).

\bibitem{xia2015effectivenessOfEuclidean}
\bibinfo{author}{Xia, S.}, \bibinfo{author}{Xiong, Z.}, \bibinfo{author}{Luo,
  Y.}, \bibinfo{author}{Zhang, G.} \emph{et~al.}
\newblock \bibinfo{journal}{\bibinfo{title}{Effectiveness of the euclidean
  distance in high dimensional spaces}}.
\newblock {\emph{\JournalTitle{Optik}}} \textbf{\bibinfo{volume}{126}},
  \bibinfo{pages}{5614--5619} (\bibinfo{year}{2015}).

\bibitem{korenius2007pcaCosineInformationRetriveal}
\bibinfo{author}{Korenius, T.}, \bibinfo{author}{Laurikkala, J.} \&
  \bibinfo{author}{Juhola, M.}
\newblock \bibinfo{journal}{\bibinfo{title}{On principal component analysis,
  cosine and euclidean measures in information retrieval}}.
\newblock {\emph{\JournalTitle{Information Sciences}}}
  \textbf{\bibinfo{volume}{177}}, \bibinfo{pages}{4893--4905}
  (\bibinfo{year}{2007}).

\bibitem{ioffe2015batchNorm}
\bibinfo{author}{Ioffe, S.} \& \bibinfo{author}{Szegedy, C.}
\newblock \bibinfo{title}{Batch normalization: Accelerating deep network
  training by reducing internal covariate shift}.
\newblock In \emph{\bibinfo{booktitle}{International conference on machine
  learning}}, \bibinfo{pages}{448--456} (\bibinfo{year}{2015}).

\bibitem{thompson2020computationalLimits}
\bibinfo{author}{Thompson, N.~C.}, \bibinfo{author}{Greenewald, K.},
  \bibinfo{author}{Lee, K.}, \bibinfo{author}{Manso, G.~F.} \emph{et~al.}
\newblock \bibinfo{journal}{\bibinfo{title}{The computational limits of deep
  learning}}.
\newblock {\emph{\JournalTitle{arXiv preprint arXiv:2007.05558}}}
  \textbf{\bibinfo{volume}{10}} (\bibinfo{year}{2020}).

\bibitem{milano2023materia}
\bibinfo{author}{Milano, G.}, \bibinfo{author}{Montano, K.} \&
  \bibinfo{author}{Ricciardi, C.}
\newblock \bibinfo{journal}{\bibinfo{title}{In materia implementation
  strategies of physical reservoir computing with memristive nanonetworks}}.
\newblock {\emph{\JournalTitle{Journal of Physics D: Applied Physics}}}
  \textbf{\bibinfo{volume}{56}}, \bibinfo{pages}{084005}
  (\bibinfo{year}{2023}).

\bibitem{fan2015hierarchical}
\bibinfo{author}{Fan, D.}, \bibinfo{author}{Sharad, M.},
  \bibinfo{author}{Sengupta, A.} \& \bibinfo{author}{Roy, K.}
\newblock \bibinfo{journal}{\bibinfo{title}{Hierarchical temporal memory based
  on spin-neurons and resistive memory for energy-efficient brain-inspired
  computing}}.
\newblock {\emph{\JournalTitle{IEEE transactions on neural networks and
  learning systems}}} \textbf{\bibinfo{volume}{27}},
  \bibinfo{pages}{1907--1919} (\bibinfo{year}{2015}).

\bibitem{demis2016nanoarchitectonic}
\bibinfo{author}{Demis, E.~C.} \emph{et~al.}
\newblock \bibinfo{journal}{\bibinfo{title}{Nanoarchitectonic atomic switch
  networks for unconventional computing}}.
\newblock {\emph{\JournalTitle{Japanese Journal of Applied Physics}}}
  \textbf{\bibinfo{volume}{55}}, \bibinfo{pages}{1102B2}
  (\bibinfo{year}{2016}).

\bibitem{drusch2012sentinelMSI}
\bibinfo{author}{Drusch, M.} \emph{et~al.}
\newblock \bibinfo{journal}{\bibinfo{title}{Sentinel-2: Esa's optical
  high-resolution mission for gmes operational services}}.
\newblock {\emph{\JournalTitle{Remote sensing of Environment}}}
  \textbf{\bibinfo{volume}{120}}, \bibinfo{pages}{25--36}
  (\bibinfo{year}{2012}).

\bibitem{knight2014landsatOLI}
\bibinfo{author}{Knight, E.~J.} \& \bibinfo{author}{Kvaran, G.}
\newblock \bibinfo{journal}{\bibinfo{title}{Landsat-8 operational land imager
  design, characterization and performance}}.
\newblock {\emph{\JournalTitle{Remote sensing}}} \textbf{\bibinfo{volume}{6}},
  \bibinfo{pages}{10286--10305} (\bibinfo{year}{2014}).

\bibitem{gorelick2017google}
\bibinfo{author}{Gorelick, N.} \emph{et~al.}
\newblock \bibinfo{journal}{\bibinfo{title}{Google earth engine:
  Planetary-scale geospatial analysis for everyone}}.
\newblock {\emph{\JournalTitle{Remote sensing of Environment}}}
  \textbf{\bibinfo{volume}{202}}, \bibinfo{pages}{18--27}
  (\bibinfo{year}{2017}).

\bibitem{Loeffler2020}
\bibinfo{author}{Loeffler, A.} \emph{et~al.}
\newblock \bibinfo{journal}{\bibinfo{title}{Topological properties of
  neuromorphic nanowire networks}}.
\newblock {\emph{\JournalTitle{Frontiers in Neuroscience}}}
  \textbf{\bibinfo{volume}{14}}, \doiprefix\url{10.3389/fnins.2020.00184}
  (\bibinfo{year}{2020}).

\bibitem{davis2006rprecisionRecall}
\bibinfo{author}{Davis, J.} \& \bibinfo{author}{Goadrich, M.}
\newblock \bibinfo{title}{The relationship between precision-recall and roc
  curves}.
\newblock In \emph{\bibinfo{booktitle}{Proceedings of the 23rd international
  conference on Machine learning}}, \bibinfo{pages}{233--240}
  (\bibinfo{year}{2006}).

\bibitem{garcia1991mapping}
\bibinfo{author}{Garc{\'\i}a, M.~L.} \& \bibinfo{author}{Caselles, V.}
\newblock \bibinfo{journal}{\bibinfo{title}{Mapping burns and natural
  reforestation using thematic mapper data}}.
\newblock {\emph{\JournalTitle{Geocarto International}}}
  \textbf{\bibinfo{volume}{6}}, \bibinfo{pages}{31--37} (\bibinfo{year}{1991}).

\bibitem{ndviCommentary}
\bibinfo{author}{Huang, S.}, \bibinfo{author}{Tang, L.}, \bibinfo{author}{Hupy,
  J.~P.}, \bibinfo{author}{Wang, Y.} \& \bibinfo{author}{Shao, G.}
\newblock \bibinfo{journal}{\bibinfo{title}{A commentary review on the use of
  normalized difference vegetation index (ndvi) in the era of popular remote
  sensing}}.
\newblock {\emph{\JournalTitle{Journal of Forestry Research}}}
  \textbf{\bibinfo{volume}{32}}, \bibinfo{pages}{1--6} (\bibinfo{year}{2021}).

\bibitem{filipponi2018bais2}
\bibinfo{author}{Filipponi, F.}
\newblock \bibinfo{title}{Bais2: Burned area index for sentinel-2}.
\newblock In \emph{\bibinfo{booktitle}{Proceedings}}, vol.~\bibinfo{volume}{2},
  \bibinfo{pages}{364} (\bibinfo{year}{2018}).

\bibitem{muhsoni2018comparison}
\bibinfo{author}{Muhsoni, F.~F.}, \bibinfo{author}{Sambah, A.~B.},
  \bibinfo{author}{Mahmudi, M.} \& \bibinfo{author}{Wiadnya, D.}
\newblock \bibinfo{journal}{\bibinfo{title}{Comparison of different vegetation
  indices for assessing mangrove density using sentinel-2 imagery}}.
\newblock {\emph{\JournalTitle{GEOMATE Journal}}}
  \textbf{\bibinfo{volume}{14}}, \bibinfo{pages}{42--51}
  (\bibinfo{year}{2018}).

\end{thebibliography}

\section*{Acknowledgements}

The authors acknowledge the use of the Artemis High Performance Computing resource at the Sydney Informatics Hub, a Core Research Facility of the University of Sydney.
S.S. is supported by an Australian Government Research Training Program (RTP) Scholarship.

\section*{Author contributions statement} 

S.S. and Z.K. designed the experiments. S.S. performed the experiments with guidance from C.P. and Z.K. S.S., C.P. and Z.K. analysed the results. S.S. drafted the manuscript with consultation from Z.K. All authors critically reviewed the manuscript. Z.K. supervised the project.

\section*{Additional information} 

\subsection*{Competing interests}

Z.K. is a founder and equity holder of Emergentia, Inc., which filed a non-provisional U.S. patent application (no. 18/334,243) for the software simulator described in this work.
C.P. was employed by Trillium Technologies during this work.
The other author declares no competing interests.

\end{document}